\documentclass{article}
\usepackage[letterpaper,margin=0.8in]{geometry}
\usepackage{graphicx} 
\usepackage{amssymb,amsfonts,amsmath,latexsym,dsfont}
\usepackage{hyperref}
\usepackage{booktabs}
\usepackage{overpic}
\usepackage{authblk}

\title{DeepSeek vs. ChatGPT vs. Claude: A Comparative Study for Scientific Computing and Scientific Machine Learning Tasks}

\author[1,*]{Qile Jiang}
\author[1,*]{Zhiwei Gao}
\author[2,†]{George Em Karniadakis}
\affil[1]{School of Engineering, Brown University, Providence, RI, USA}
\affil[2]{Division of Applied Mathematics, Brown University, Providence, RI, USA}

\date{February 2025}

\begin{document}

\maketitle
\let\thefootnote\relax\footnotetext{* These authors contributed equally.}
\footnotetext{† Corresponding author: \texttt{george\_karniadakis@brown.edu}}

\begin{abstract}
    Large Language Models (LLMs) have emerged as powerful tools for tackling a wide range of problems, including those in scientific computing, particularly in solving partial differential equations (PDEs). However, different models exhibit distinct strengths and preferences, resulting in varying levels of performance. In this paper, we compare the capabilities of the most advanced LLMs—DeepSeek, ChatGPT, and Claude—along with their reasoning-optimized versions in addressing computational challenges. Specifically, we evaluate their proficiency in solving traditional numerical problems in scientific computing as well as leveraging scientific machine learning techniques for PDE-based problems. We designed all our experiments so that a non-trivial decision is required, e.g. defining the proper space of input functions for neural operator learning. Our findings show that reasoning and hybrid-reasoning models consistently and significantly outperform non-reasoning ones in solving challenging problems, with ChatGPT o3-mini-high generally offering the fastest reasoning speed. 
\end{abstract}

\section{Introduction}

Large language models (LLMs), such as ChatGPT \cite{openai_gpt_4o_2024, openai_o3_mini_2025, openai_o3_mini_system_card_2024}, Claude \cite{anthropicclaude3.7sonnet}, and, most recently, DeepSeek \cite{deepseekai2024deepseekv3technicalreport} are general-purpose artificial intelligence (AI) chatbots based on the transformer architecture \cite{waswani2017attention}, trained on large amounts of data from the Internet, and have learned to engage in conversations based on user requests. Among their many applications, there has been growing interest in leveraging LLMs for scientific research, particularly in coding, mathematics, and problem-solving. This interest has driven rapid recent advancements in LLM development. For example, on January 31, 2025, OpenAI released ChatGPT o3-mini \cite{openai_o3_mini_2025}, the latest reasoning model supposed to deliver particular strength in science, math, and coding. This came only a few days after DeepSeek unveiled its own reasoning model DeepSeek-R1 \cite{deepseekai2025deepseekr1incentivizingreasoningcapability} that also specializes in math, coding, and reasoning tasks. Most recently, on February 24, 2025, Anthropic also announced  Claude 3.7 Sonnet with extended thinking mode to improve its performance on math, physics, instruction-following, and coding \cite{anthropicclaude3.7sonnet}.
These models build on prior advancements such as ChatGPT-4o \cite{openai_gpt_4o_2024}, DeepSeek V3 \cite{deepseekai2024deepseekv3technicalreport}, and Claude 3.5 Sonnet \cite{anthropicclaude3.5sonnet} marking a new phase of competition among LLM developers seeking to optimize performance in scientific fields.

LLMs have already demonstrated potential in various domains of scientific research, not only in synthesizing information \cite{antu2023using, li2024chatcite} and as a coding assistant \cite{nam2024using, liu2024exploring, chew2023llm}, but also being utilized to solve complex domain-specific problems, including material sciences \cite{zhang2024honeycomb, hong2023chatgpt, cheng2023challenges}, genetics \cite{chatterjee2023can, mcgrath2024comparative}, medical imaging \cite{srivastav2023chatgpt, hu2024advancing, yang2023impact}, and computational fluid dynamics \cite{chen2024metaopenfoam, sawada2011llm, herzog2002llm}. The integration of LLMs into scientific workflows has expanded their role from mere information retrieval tools to automated agents capable of performing reasoning-based tasks to aid and complement human researchers. However, despite these promising applications, LLMs can still exhibit fundamental limitations that raise concerns about their reliability in scientific research and computing \cite{hadi2023survey, rossi2024problems, rane2023contribution, pal2024ai, kunz2024properties}. In particular, when tasked to solve complex scientific problems, they may produce hallucinated reasoning \cite{li2023deceptive}, low level of mathematical cognition \cite{evans2024evaluating}, and even self-contradiction \cite{mundler2023self}. Unlike conversational AI tasks, where fluency and coherence are primary evaluation criteria, scientific and engineering problems demand precision,  logical consistency, and rigorous reasoning. The limitations of LLMs have thus made them unsuitable for high-stakes applications. 
 
Given these challenges, researchers have designed experiments for assessing LLM's capabilities in scientific and computational tasks, for example, benchmarking LLM's skills in mathematical reasoning \cite{liu2024mathbench, chernyshev2024u}, scientific reviews \cite{wu2023gpt, cai2024sciassess}, and engineering documentation \cite{doris2025designqa}. Rather than relying solely on generic language model benchmarks, domain-specific model testing and evaluations are essential for assessing LLMs not only for correctness but also for reasoning depth, reliability, and their ability to generalize to research-level scientific problems. 

With the intensifying competition among LLM developers, this study conducts a comparative testing of their models in computational mathematics and scientific machine learning. Deliberately tricky problems are formulated and posed  to both reasoning and non-reasoning models from the DeepSeek, ChatGPT, and Claude LLM families. We first test the LLMs' capability in numerical methods, including numerical integration, finite difference methods (FDM), and finite element methods (FEM). Since machine learning methods have gained popularity as alternatives to traditional numerical approaches \cite{thiyagalingam2022scientific, karniadakis2021physics}, we also assess the LLMs' performance in scientific machine learning tasks, including image recognition, physics-informed learning with physics-informed neural networks (PINNs) \cite{raissi2019physics}, and operator learning methods such as Deep Operator Network (DeepONet) \cite{deeponet}. By comparing the models' performances and their reasoning decisions across a variety of benchmark problems, the work provides insight into how well these popular LLMs perform in computational mathematics and the risks and benefits of adopting them in our studies and research.

\section{Experiments}

In this section, we evaluate six different LLMs developed by DeepSeek, OpenAI, and Anthropic by testing them on a range of challenging benchmark problems in numerical algorithms and scientific machine learning. The models under consideration are:

\begin{enumerate}
    \item DeepSeek V3 \cite{deepseekai2024deepseekv3technicalreport}: A general-purpose model developed by DeepSeek, trained on a broad dataset covering multiple domains. While it is not explicitly specialized for scientific tasks, its training includes extensive exposure to mathematical and engineering-related data. 
    \item DeepSeek R1 \cite{deepseekai2025deepseekr1incentivizingreasoningcapability}: A model developed by DeepSeek specifically designed for reasoning tasks, including mathematics and coding. Compared to DeepSeek V3, R1 is fine-tuned with additional reinforcement learning strategies to improve logical consistency and structured problem-solving. It is positioned as a direct competitor to OpenAI’s reasoning models.  
    \item ChatGPT 4o \cite{openai_gpt_4o_2024}: OpenAI’s current flagship model designed for general reasoning and multimodal capabilities, including text, code, and vision processing. It employs an optimized transformer architecture and focuses on improving efficiency and reducing latency while maintaining performance in coding, mathematics, and logical inference.
    \item ChatGPT o3-mini-high \cite{openai_o3_mini_2025}: ChatGPT o3-mini-high is a variant of OpenAI's o3-mini model, optimized for tasks requiring intensive reasoning in coding, mathematics, and science. ``High" stands for higher-intelligence. Despite its smaller size, o3-mini-high is supposed to deliver performance comparable to larger models by employing advanced reasoning techniques.
    \item Claude 3.7 Sonnet \cite{anthropicclaude3.7sonnet}: Anthropic's high-performance model designed for versatile text processing, code generation, and analytical tasks. It employs an optimized transformer architecture with enhanced contextual understanding capabilities and is supposed to deliver reliable performance across various domains including creative writing, programming, and knowledge-intensive applications. It claims to be the first ``hybrid reasoning model" on the market. 
    \item Claude 3.7 Sonnet extended thinking \cite{anthropicclaude3.7sonnet}: An enhanced variant of Anthropic's Sonnet model featuring an additional reasoning layer that enables deeper analytical thinking. This extended version incorporates a specialized processing mode that allows for more thorough evaluation of complex problems to improve performance on tasks requiring multi-step reasoning, mathematical problem-solving, and logical deduction while maintaining the base model's versatility.

\end{enumerate}

To ensure fairness, all chat memory and user-personalized settings are disabled. This prevents models from benefiting from prior context and ensures that each query is treated independently. Additionally, since LLM response times can be affected by server latency rather than actual evaluation speed, we do not compare or report response times for non-reasoning models (including the hybrid reasoning Claude 3.7 Sonnet). Instead, we only present in this paper the self-reported reasoning times for reasoning LLMs. The focus of our evaluation is on the quality of the generated solutions, and, for reasoning models, their reasoning process and decision-making. To differentiate the model performances, the test problems selected are tricky but also leave enough flexibility for the LLMs to make their own decisions in what method and parameters to use in their solutions. Most problems are advanced, typically at the PhD level, and demand a deep understanding of computational mathematics and familiarity with recent research advancements. The coding languages tested are chosen to be TensorFlow \cite{tensorflow2015-whitepaper} and Jax \cite{jax2018github} in Python, which are frameworks frequently used in the scientific machine learning community. 

This section will be divided into two parts. In the first, we conduct experiments to test the models' knowledge in implementing traditional numerical methods in applied mathematics, particularly in solving ordinary and partial differential equations (PDEs), such as finite difference and finite element methods. In the second part, we test the models on scientific machine learning tasks, including learning digits from the MNIST dataset and implementing PINNs and DeepONets for different problems. 

\subsection{Traditional numerical methods}

In this section, we present the six LLMs under consideration addressing a variety of problems in traditional numerical methods and solving differential equations. We compare the performance of their generated solutions and discuss the decisions the LLMs made. 

\subsubsection{Numerical solution to stiff ODEs}

The Robertson problem \autoref{Robertson ODE} is a well-known example of a stiff system of ODEs that describes the chemical reaction of H. H. Robertson \cite{robertson1966solution}: 

\begin{align}
    \frac{dx}{dt} & = -0.04x + 10^4 yz, \nonumber \\ 
\frac{dy}{dt} & = 0.04x - 10^4 yz - 3 \times 10^7 y^2, \label{Robertson ODE} \\
\frac{dz}{dt} & = 3 \times 10^7 y^2. \nonumber
\end{align}

The problem models the concentrations of three species $x(t), y(t), z(t)$ over time with initial conditions $x=1, y=z=0$. The reactions have very different time scales, particularly for $y(t)$. Solving such stiff ODEs with explicit integration methods leads to unstable behavior unless an extremely small step size is used. Implicit methods such as implicit Runge–Kutta and backward differentiation formula (BDF) are usually more suited to solving these problems. 

To assess if the LLMs can implement an appropriate numerical scheme for solving this very stiff system, we ask them with the following prompt:

\texttt{Implement an appropriate numerical method from scratch to solve the following system of ODEs  \\ in Python: 
$$
\frac{dx}{dt} = -0.04x + 10^4 yz, \quad 
\frac{dy}{dt} = 0.04x - 10^4 yz - 3 \times 10^7 y^2,\quad
\frac{dz}{dt} = 3 \times 10^7 y^2. \quad
$$
on a time interval of $t \in [0, 500]$ with the initial conditions $x=1, y=z=0$. }

\begin{figure}[h!]
    \centering
    \includegraphics[width=1\linewidth]{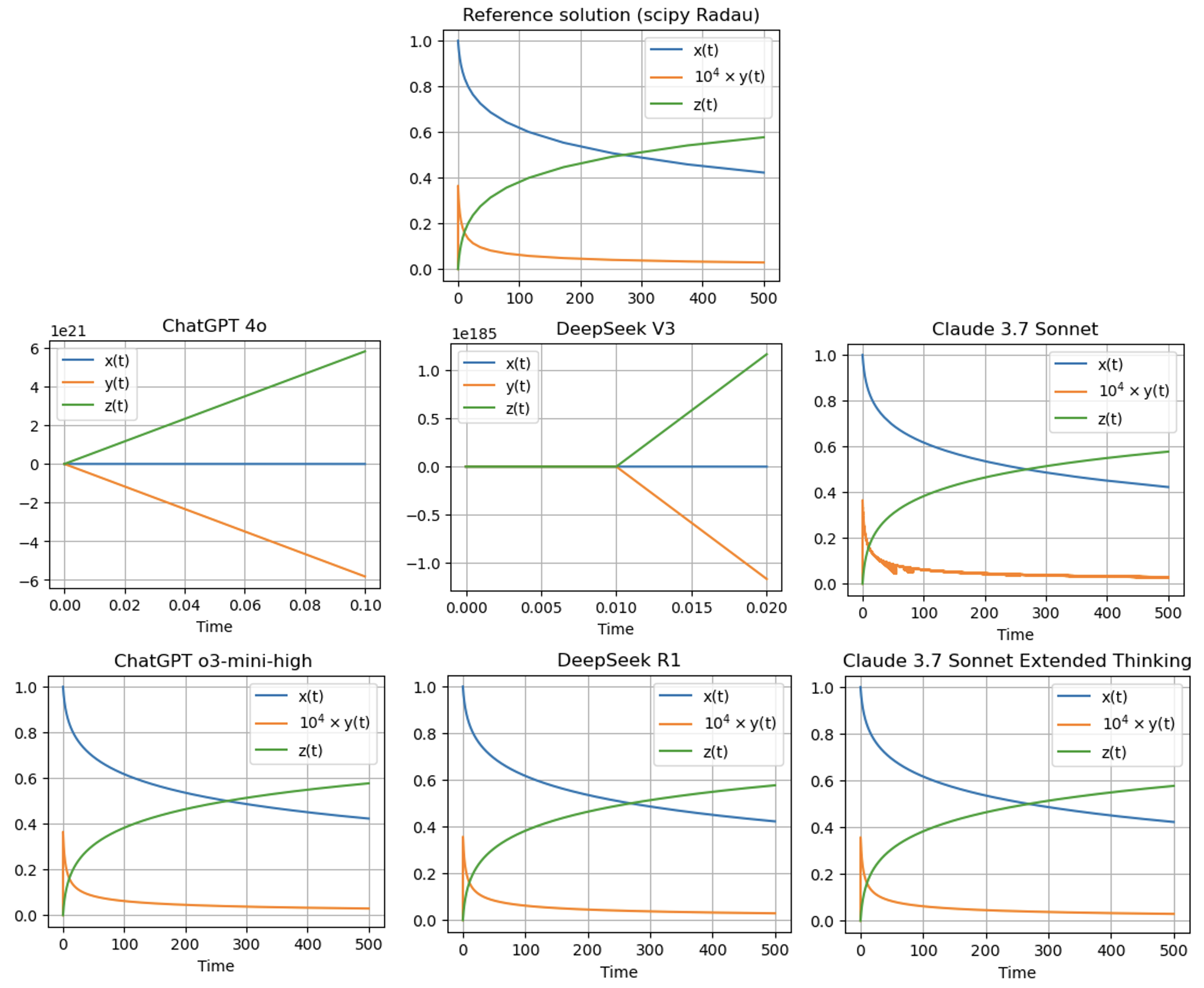}
    \caption{Solutions to the Robertson ODEs computed using the numerical scheme selected by each LLM. For reference, the ODE system is also solved in \texttt{scipy} using the Radau method. For better visualization, the concentration for the $y$ species is scaled by $10^4$ in plots where the solution converges.}
    \label{fig:ODE_solution}
\end{figure}

The LLMs' responses are summarized in \autoref{tab:ODE_responses}, and the computed solutions are plotted in \autoref{fig:ODE_solution}.  For reference, the ODE system is also solved in \texttt{scipy} using an implicit Runge-Kutta method of Radau IIA family of order 5 (Radau) \cite{hairer1991ii}.

\begin{table}[h!]
    \centering
    \renewcommand{\arraystretch}{1.5}
    \caption{Numerical methods and parameters chosen by each LLM to solve the system of Robertson ODEs. The error represents the relative difference between the LLM's solution and the reference solution from \texttt{scipy}'s Radau method implementation.}
    \vspace{2mm}
    \resizebox{0.99\linewidth}{!}{%
    \begin{tabular}{c|c|c|c|c|c|c}
    \toprule
    Model &  \begin{tabular}{@{}l@{}} Reasoning \\ time (s) \end{tabular}  &Method & Step Size & $L_2$ Error ($x$) & $L_2$ Error ($y$) & $L_2$ Error ($z$) \\ \hline
    DeepSeek V3 &  N/A&RK4 & 0.01 & N/A & N/A & N/A \\  \hline
    DeepSeek R1 &  249.0&Backward Euler & 0.1 & \textbf{$7.86 \times 10^{-6} \%$} & \textbf{$7.61 \times 10^{-5} \%$} & \textbf{$1.96 \times 10^{-4} \%$} \\  \hline
    ChatGPT 4o &  N/A&RK4 & 0.1 & N/A & N/A & N/A \\ \hline
    ChatGPT &  88.0&Backward Euler with& Adaptive  & $3.93 \times 10^{-6} \%$ & $9.40 \times 10^{-4} \%$ & $0.14 \%$ \\ 
     o3-mini-high&  &adaptive time stepping & ($10^{-6}$ initially) &  &  & \\ \hline
    Claude 3.7 Sonnet &  N/A&RK4 with & Adaptive  & $\mathbf{1.24 \times 10^{-6} \%}$& $0.05 \%$& $4.89 \times 10^{-4} \%$\\ 
     &  &adaptive time stepping & ($10^{-4}$ initially)&  &  & \\ \hline
     \begin{tabular}{@{}l@{}}Claude 3.7 Sonnet \\ extended thinking\end{tabular}  &  115.0&Backward Euler & $0.1$& $7.74\times 10^{-6} \%$& $\mathbf{7.53 \times 10^{-5} \%}$& $\mathbf{1.94 \times 10^{-4}} \%$\\ 
    \bottomrule
    \end{tabular}
    }
    \label{tab:ODE_responses}
\end{table}

All non-reasoning models implemented a fourth-order Runge-Kutta (RK4) scheme. The versions by DeepSeek V3 and ChatGPT 4o chose relatively large step sizes, and they failed to solve the problem and yielded exponentially growing solutions. In contrast, the hybrid-reasoning Claude 3.7 Sonnet implemented an RK4 scheme with adaptive time stepping that compares one full step of size $\Delta t$ with two half‑steps (each of size $\Delta t /2$) to estimate the local truncation error. If the estimated error is below a prescribed tolerance, the step is accepted; otherwise, it is rejected and a new (smaller) time step is computed. Claude 3.7 Sonnet's adaptive method yielded much better accuracy than the other two non-reasoning models, although errors on the $y$-species are still visible. On the other hand, all fully reasoning models implemented the backward Euler method. In particular, ChatGPT o3-mini-high used a similar adaptive time stepping method. Overall, Claude 3.7 Sonnet with extended thinking achieved the best results. ChatGPT o3-mini-high yielded satisfactory solutions with the shortest reasoning time. 

We remark that all three reasoning models—DeepSeek R1,  ChatGPT o3-mini-high, and Claude 3.7 Sonnet with extended thinking—recognized the stiffness of the ODE system and chose an implicit method, while all non-reasoning models failed to do so. During the reasoning process, DeepSeek R1 reported that 

\texttt{"Given the time interval is up to 500, and the equations have terms with coefficients like \\ 1e4 and 3e7, which are quite large, this might be a stiff system. If the system is stiff, using \\ an implicit method like backward Euler or a solver designed for stiff systems (like Rosenbrock \\ methods) would be better."}

ChatGPT o3-mini-high reported that 

\texttt{"Since it's stiff, explicit methods like Euler or RK4 won't work well. I think we'll need \\ something like an implicit method, maybe implicit Euler, or a modified explicit method like an \\ adaptive Runge-Kutta."}

Similarly, Claude 3.7 Sonnet with extended thinking reported that 

\texttt{"This is a stiff system of ODEs, characterized by the vastly different coefficients \ldots I'll \\ implement the Backward Differentiation Formula (BDF) method, which is specifically designed for \\ stiff problems." }

This experiment demonstrates the superior performance of the reasoning LLMs over non-reasoning ones, as reasoning models are able to analyze the problem first before choosing an appropriate method to implement, resembling the approach of a human scientist.

\subsubsection{Finite difference for Poisson equation}

In this experiment, we test the capability of DeepSeek, ChatGPT, and Claude models to implement finite difference methods for partial differential equations. We consider the following Poisson equation \cite{buzbee1970direct}:
\begin{equation*}
    \left\{ \begin{array}{l}
          -\Delta u = f, \quad x \in \Omega  \\
          u|_{\partial \Omega} = 0,
    \end{array}
    \right.
\end{equation*}
and set $\Omega = [-1,1]^{2}/[0,1]^{2}$ to be an L-shaped domain. The right hand side function is set to be $f(x, y) = 1$. We are specifically interested in testing if the LLMs can solve equations on a non-standard domain. The following question was asked to all of LLMs:

\texttt{Use finite difference method to solve the 2D Poisson equation in L-shaped domain $[-1,1]^{2}/[0, 1]^{2}$ with zero boundary condition and right hand side $f(x,y) = 1$, where the coding language is \texttt{Python}}. 

\begin{figure}[h!]
    \centering
    \begin{overpic}[width = 0.4\textwidth]{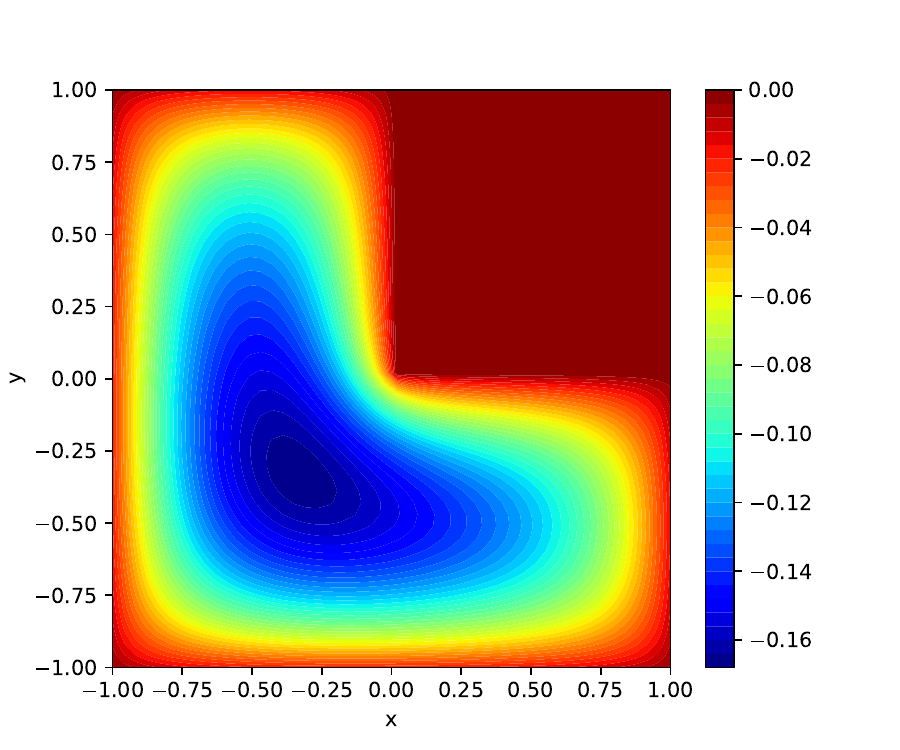}
    \put(25,75) {ChatGPT 4o}
    \end{overpic}
    \begin{overpic}[width = 0.4\textwidth]{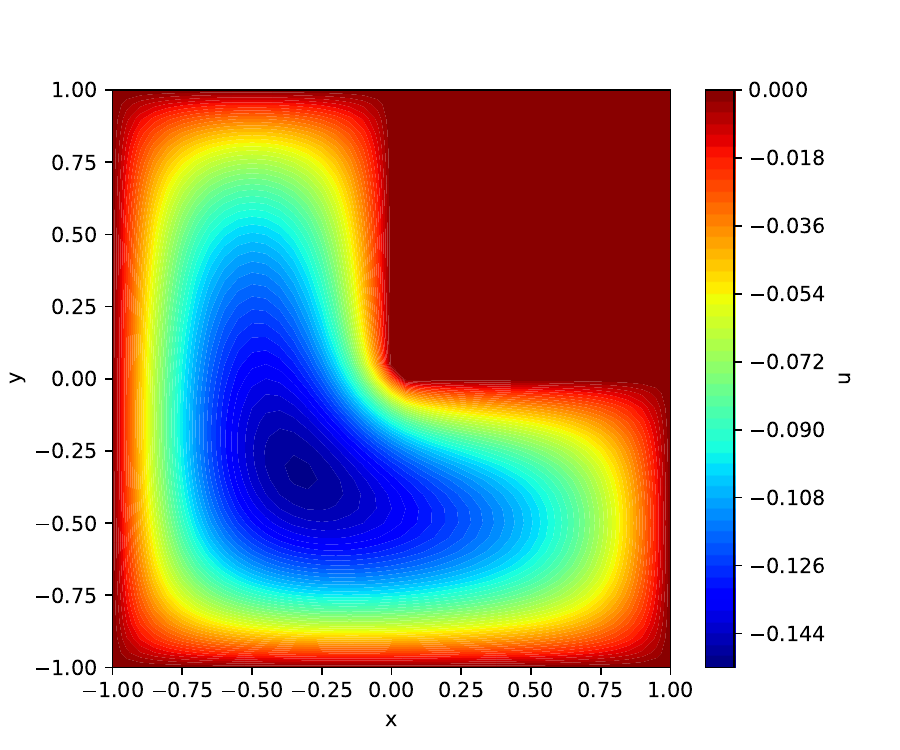}
    \put(15,75) {ChatGPT o3-mini-high}
    \end{overpic}\\ 
    \begin{overpic}[width = 0.41\textwidth]{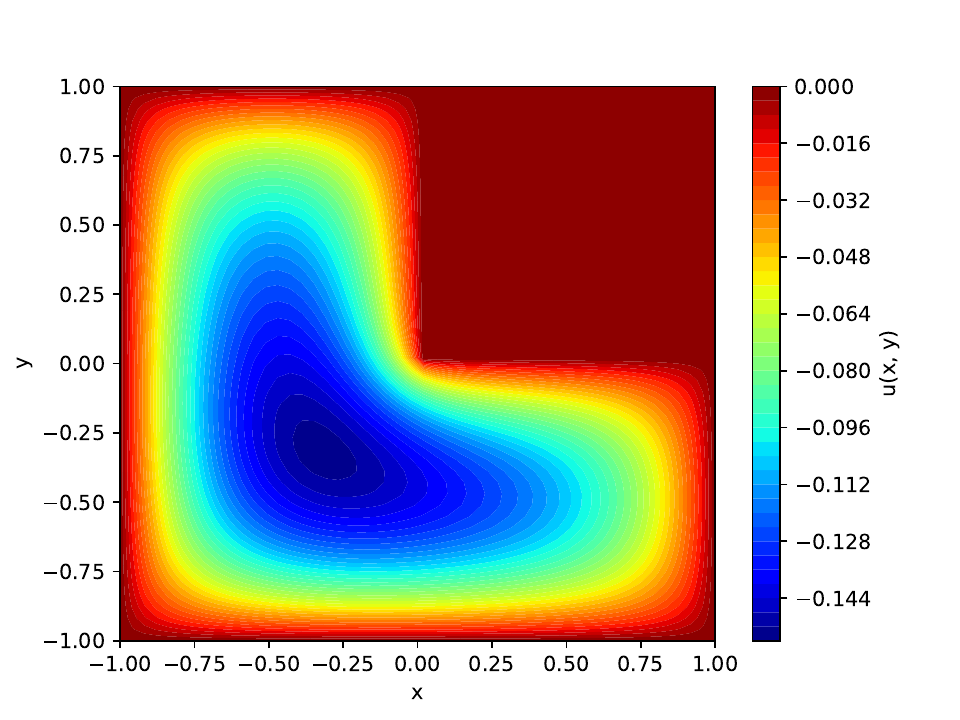}
    \put(28,68) {DeepSeek V3}
    \end{overpic}
    \hspace{-0.4cm}
    \begin{overpic}[width = 0.42\textwidth]{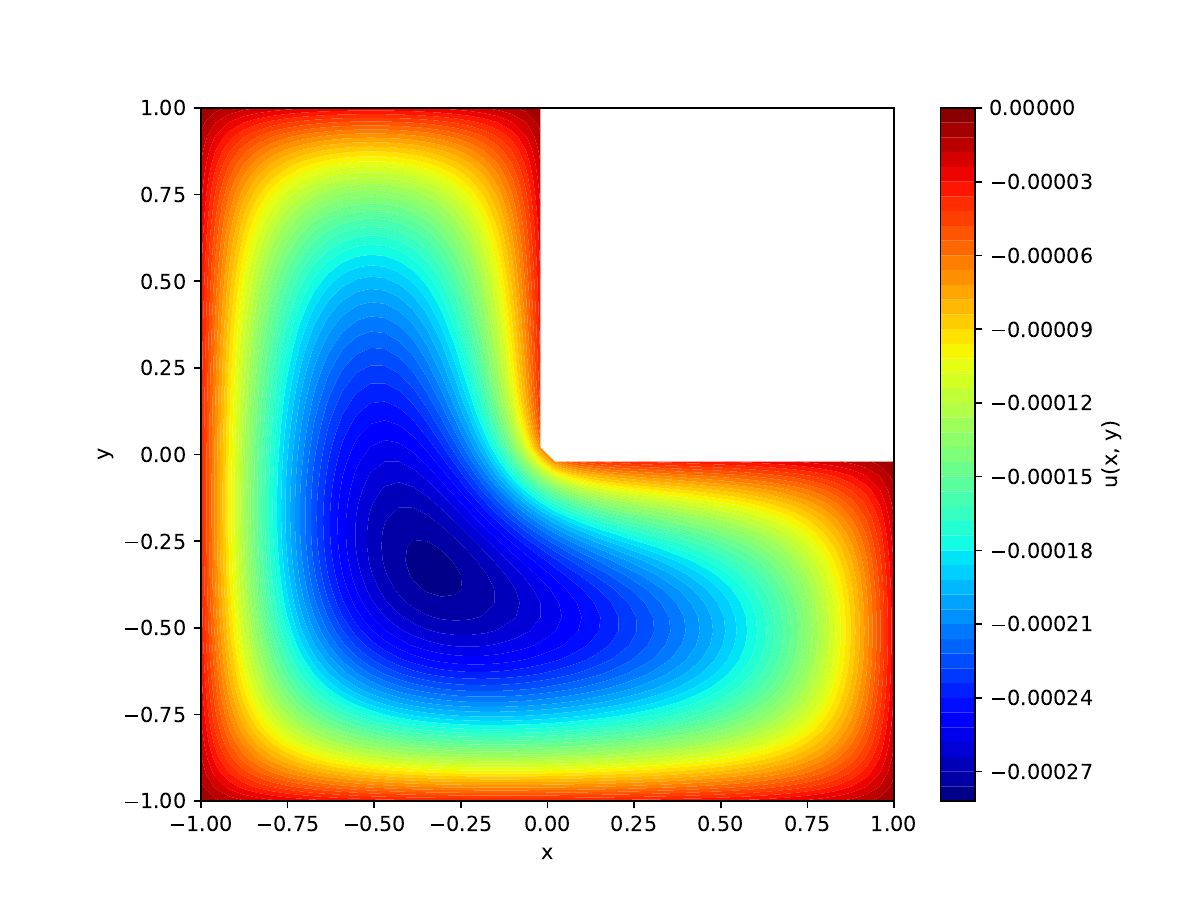}
    \put(28,68) {DeepSeek R1}
    \end{overpic}
    \begin{overpic}[width = 0.43\textwidth]{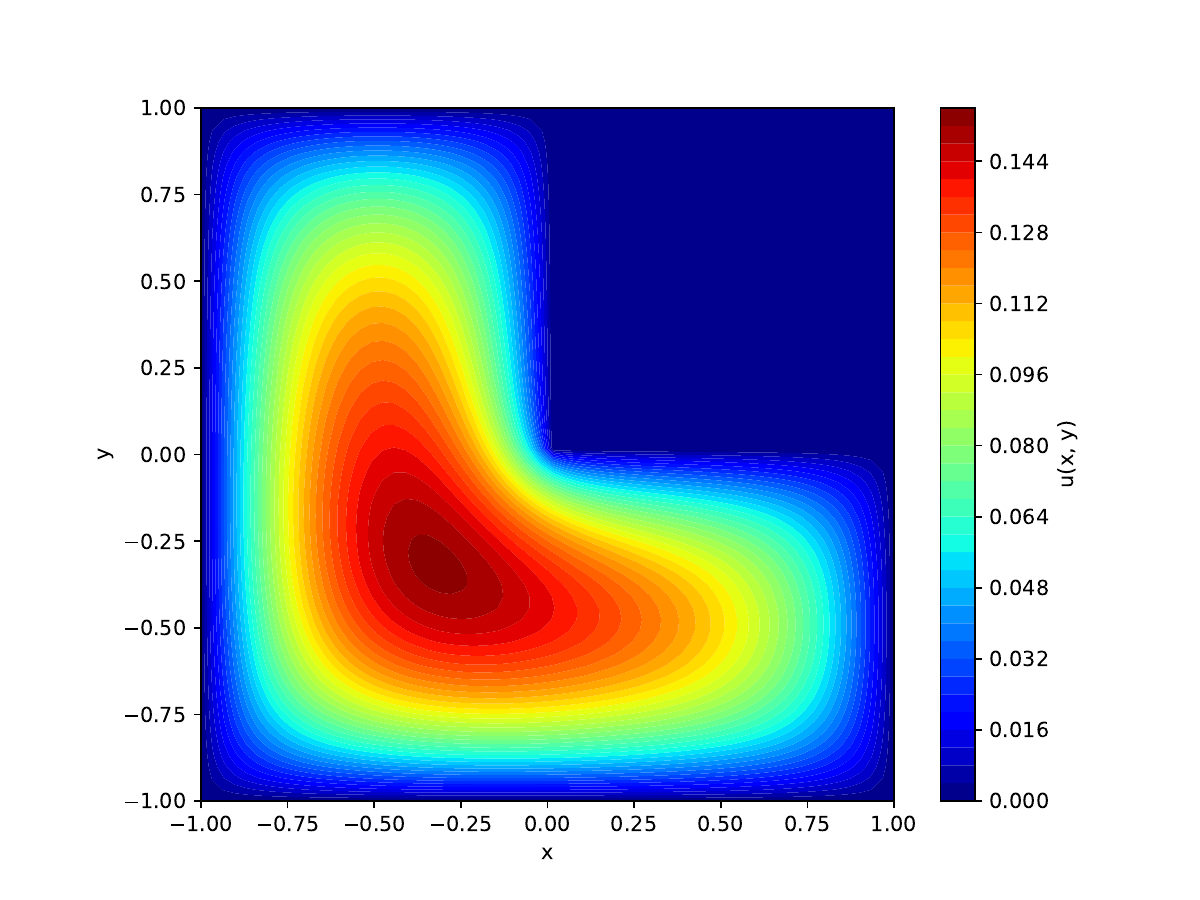}
    \put(30,68) {Claude 3.7 Sonnet}
    \end{overpic}
    \hspace{-0.4cm}
    \begin{overpic}[width = 0.43\textwidth]{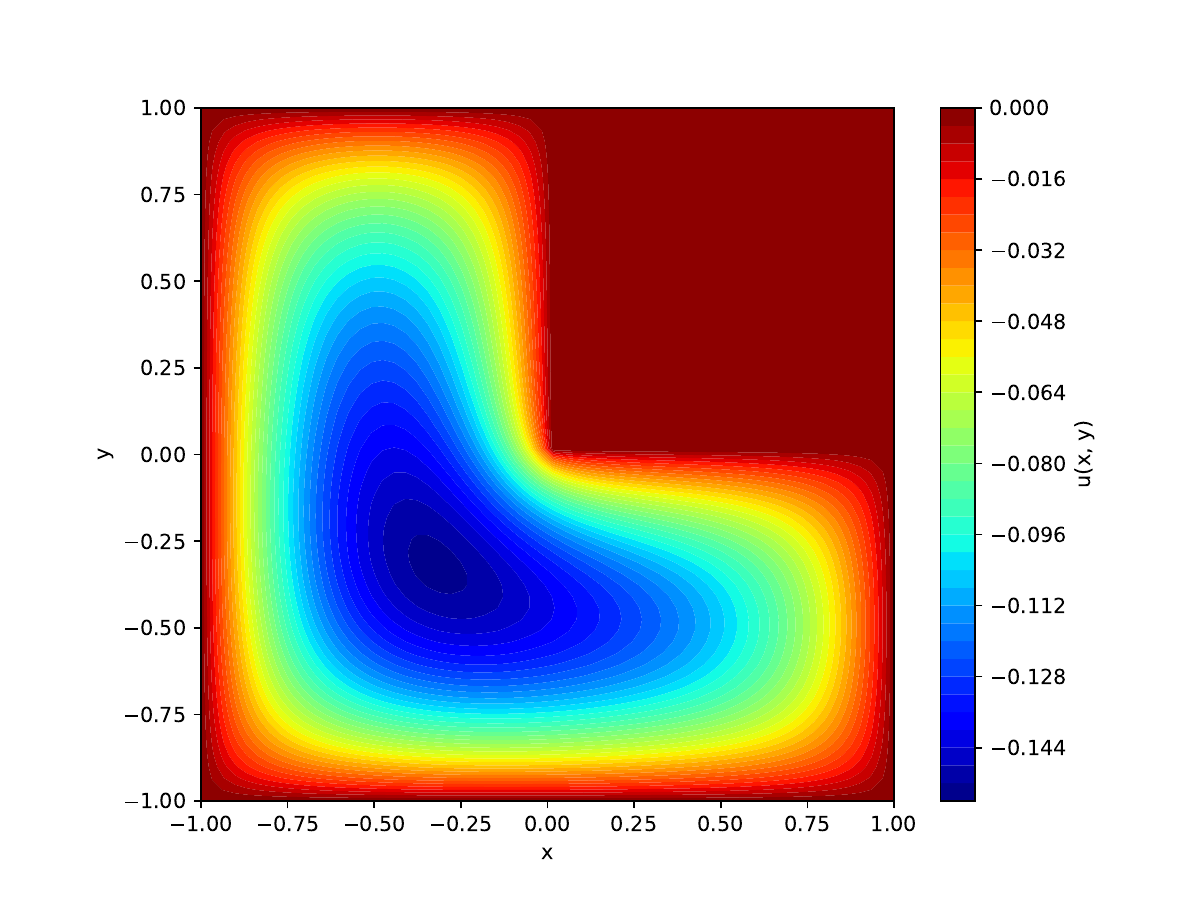}
    \hspace{-1cm} \put(26,68) {Claude 3.7 Sonnet extended thinking}
    \end{overpic}
    \caption{The predicted solutions given by different models for the Poisson equation.}
    \label{fig:poisson_solution}
\end{figure}

Except for ChatGPT, all of other models provided error-free, executable codes. Notably, DeepSeek employed the most advanced iterative methods to solve the problem, whereas the others relied on conventional finite difference methods for discretization and solving the linear system. The predicted solutions are plotted in \autoref{fig:poisson_solution}.

Results and methods used by each LLM are listed in \autoref{tab:possion}. Among all six models, the only correct result is given by Claude 3.7 Sonnet.  DeepSeek V3's implementation takes a much longer CPU time to give the results due to its using iteration methods. DeepSeek-R1 not only mistakes the sign of the solution but gives wrong scales, leading to worse performance. All other models, except the correct Claude 3.7 Sonnet, only mistake the sign of the solution. Furthermore, among the three reasoning models, ChatGPT o3-mini-high responded much faster to requests than DeepSeek R1 and Claude 3.7 Sonnet with extended thinking.

\begin{table}[h!]
    \centering
    \renewcommand{\arraystretch}{1.5}
    \caption{Summary of performance of different models in solving the Poisson equation.}
    \vspace{2mm}
    \begin{tabular}{c|c|c|c|c|c}
    \toprule 
         &Reasoning time(s)& Method & Grid & CPU time (s) & 
         $L_{2}$ error \\
         \hline
        DeepSeek V3 & N/A& Iteration & $50\times 50$ & 9.5 & 201\%\\ 
        \hline
        DeepSeek R1 &476.0 &FDM& $50\times 50$ & 0.6& 203\%\\ 
        \hline
        ChatGPT 4o &N/A&FDM& $50\times 50$ & 0.6 & 203\%\\
        \hline
        ChatGPT o3-mini-high &32.0&FDM& $41\times 41$ & 0.5 & 207\%\\ 
        \hline
        Claude 3.7 Sonnet&N/A& FDM&$50\times 50$ & 0.6& \textbf{4.54\%}\\
        \hline
        Claude 3.7 Sonnet extended thinking&125.0& FDM&$60\times 60$ & 0.7&202\% \\ 
        \bottomrule
    \end{tabular}
    \label{tab:possion}
\end{table}

\subsubsection{Finite element methods for Vibrating Beam equation}
\begin{figure}[h!]
    \centering
    \includegraphics[width=0.5\linewidth]{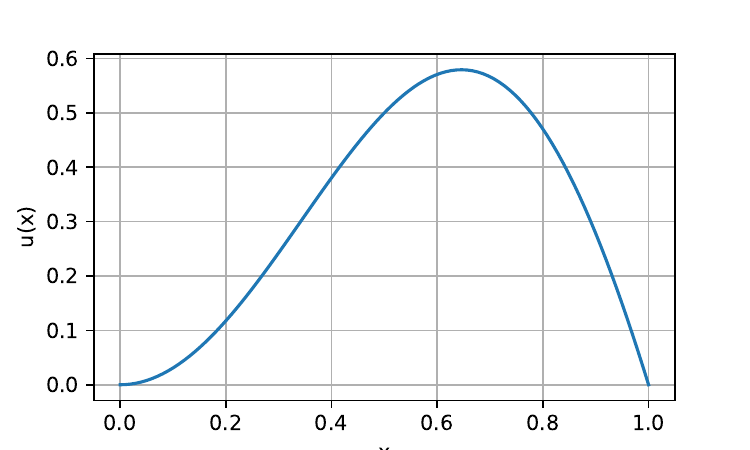}
    \caption{The true solution for the beam equation.}
    \label{fig:beam_true_solution}
\end{figure}

In this experiment, we consider the beam equation \cite{eliasson2016kam}:
\begin{align*}
EI \frac{d^4 u}{dx^4} = \pi^4 x \sin \pi x - 4\pi^3 \cos \pi x, \quad 0 \leq x \leq L,
\end{align*}
with corresponding non-homogeneous boundary conditions:
\begin{align*}
&u(0) = 0, \quad u''(0) = 2\pi,\\
&u(L) = 0, \quad u''(L) = -2\pi. 
\end{align*}
Here, $EI = 1N/m^{2}$ is the constant flexural rigidity. For simplicity, we set $L = 1$. The true solution, plotted in \autoref{fig:beam_true_solution}, is obtained by solving the ODE and given as 
\begin{equation*}
    u(x) = x\sin(\pi x).
\end{equation*}

All LLMs under consideration are asked with the following question: 

\texttt{Using finite element method from scratch to solve the beam equation:
\begin{equation*}
\left\{\begin{array}{l}
    \frac{d^4 u}{dx^4} = \pi^4 x \sin \pi x - 4\pi^3 \cos \pi x, \quad 0 \leq x \leq 1,\\ 
    u(0) = 0, \quad u''(0) = 2\pi,\\ 
    u(1) = 0, \quad u''(1) = -2\pi.
\end{array}\right.
\end{equation*}
The coding language is Python.
}

Results and details of the implementation can be found in \autoref{tab:beam}. The solutions predicted by each LLM are plotted in \autoref{fig:fem_solution}. Our results demonstrate that none of the LLMs can correctly apply finite element methods to get a very accurate solution.

\begin{table}[h!]
    \centering
    \renewcommand{\arraystretch}{1.5}
    \caption{Summary of performance of different models in solving the beam equation.}
    \vspace{2mm}
    \begin{tabular}{c|c|c|c|c}
    \toprule
   &Reasoning time (s)& Basis & Grid & $L_{2}$ error\\ 
    \hline 
    DeepSeek V3     &N/A& Hermite &100 & 276300\% \\
    \hline 
    DeepSeek R1     &790.0&Hermite & 10 & 15.3\%  \\ 
    \hline 
    ChatGPT 4o    &N/A& Hermite & 50 & 205\% \\ 
    \hline 
    ChatGPT o3-mini-high &31.0 & Hermite & 100 & 19.4\% \\
    \hline 
    Claude 3.7 Sonnet &N/A&Hermite & 100 & 957\% \\ 
    \hline 
    Claude 3.7 Sonnet extended thinking &9.0 & Hermite & 20 & \textbf{13.7\%}\\ 
    \bottomrule
    \end{tabular}
    \label{tab:beam}
\end{table}

 We observe that all LLMs use Hermite basis functions to approximate the solution. Besides, they realize that the weak form should be derived and then the Galerkin methods can be applied. However, they all choose the wrong test space to derive the weak form, which leads to the final wrong results. Among these models, the non-reasoning models perform much worse than reasoning models, showing significantly larger $L_2$ errors. In contrast, consistent with results from other experiments, the reasoning models solve the problem more effectively, with Claude 3.7 Sonnet extended thinking achieving the smallest $L_2$ error. For response times among reasoning models, ChatGPT o3-mini-high and Claude 3.7 Sonnet with extended thinking responded much faster than DeepSeek R1.

Moreover, the implementations by Claude will further test convergence rates of the method using different numbers of basis functions. It turns out that Claude Sonnet models reason more technically like human scientists for mathematical problems and give comprehensive results to verify which method is better.

\begin{figure}[h!]
    \centering
    \begin{overpic}[width = 0.45\textwidth]{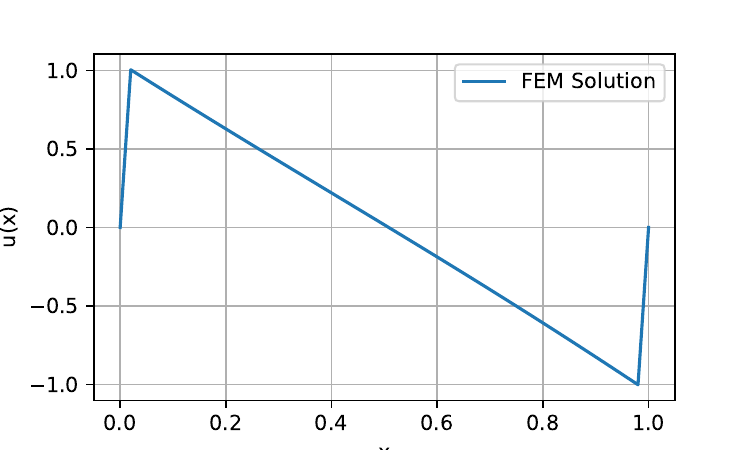}
    \put(35,55) {ChatGPT 4o}
    \end{overpic}
    \begin{overpic}[width = 0.45\textwidth]{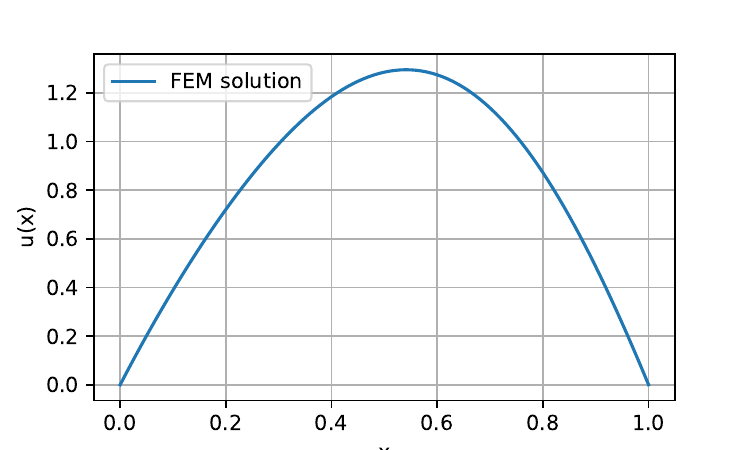}
    \put(30,55) {ChatGPT o3-mini-high}
    \end{overpic}
    \begin{overpic}[width = 0.45\textwidth]{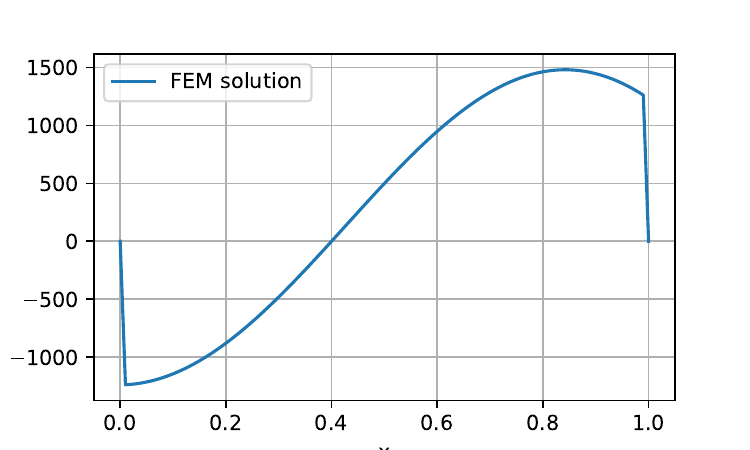}
    \put(35,55) {DeepSeek V3}
    \end{overpic}
    \begin{overpic}[width = 0.45\textwidth]{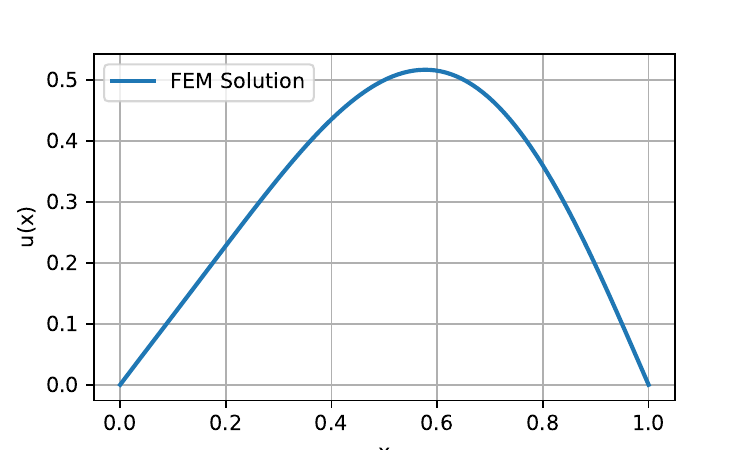}
    \put(35,55) {DeepSeek R1}
    \end{overpic}
    \begin{overpic}[width = 0.45\textwidth]{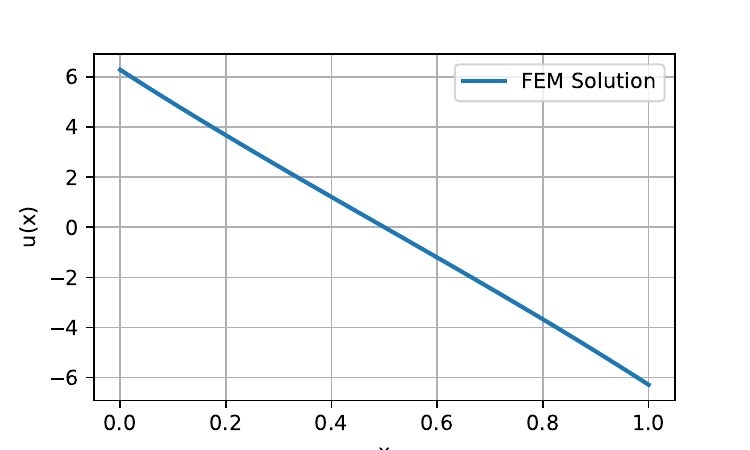}
    \put(35,55) {Cluade 3.7 Sonnet}
    \end{overpic}
    \begin{overpic}[width = 0.45\textwidth]{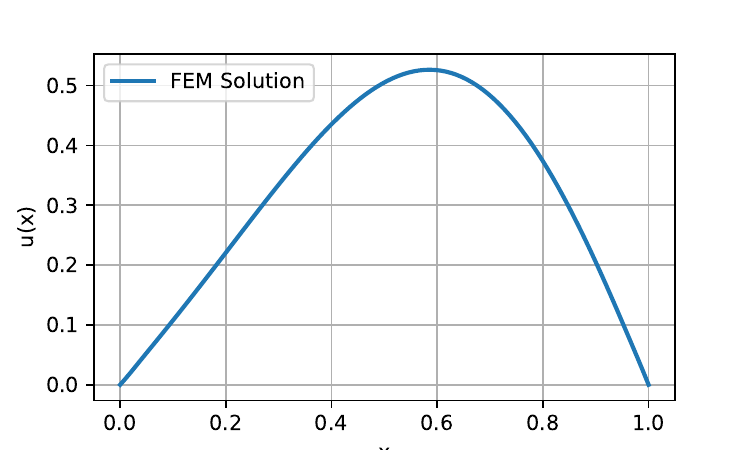}
    \hspace{-1cm} \put(30,55) {Claude 3.7 Sonnet extended thinking}
    \end{overpic}
    \caption{Predicted solutions from different LLMs for the beam equation.}
    \label{fig:fem_solution}
\end{figure}

\subsubsection{Quadratures for integrals}
Next, we evaluate the LLMs' ability to apply quadrature rules correctly for numerically computing integrals with singularities. Consider the following integral:
\begin{equation*}
    I = \int_{0}^{1}\frac{e^{-x}}{x^{\frac{2}{3}}}dx.
\end{equation*}

The question posed to all of the models is

\texttt{Using quadrature rules to calculate the integral 
$\int_{0}^{1}\frac{e^{-x}}{x^{\frac{2}{3}}}dx$ with python from scratch.
}

All LLMs provided error-free implementation, while the approaches of different models vary significantly. DeepSeek V3 directly used the \texttt{scipy.integrate} module to compute the integral, while the other models first applied a transformation to remove the singularity. Specifically, setting $x = t^{3}$ gives
\begin{equation*}
    I = \int_{0}^{1} e^{-t^3} \cdot \frac{3t^2}{(t^3)^{2/3}} dt
= \int_{0}^{1} 3e^{-t^3} t^{-2+2} dt
= 3 \int_{0}^{1} e^{-t^3} dt.
\end{equation*}

\begin{figure}
    \centering
\includegraphics[width=0.5\linewidth]{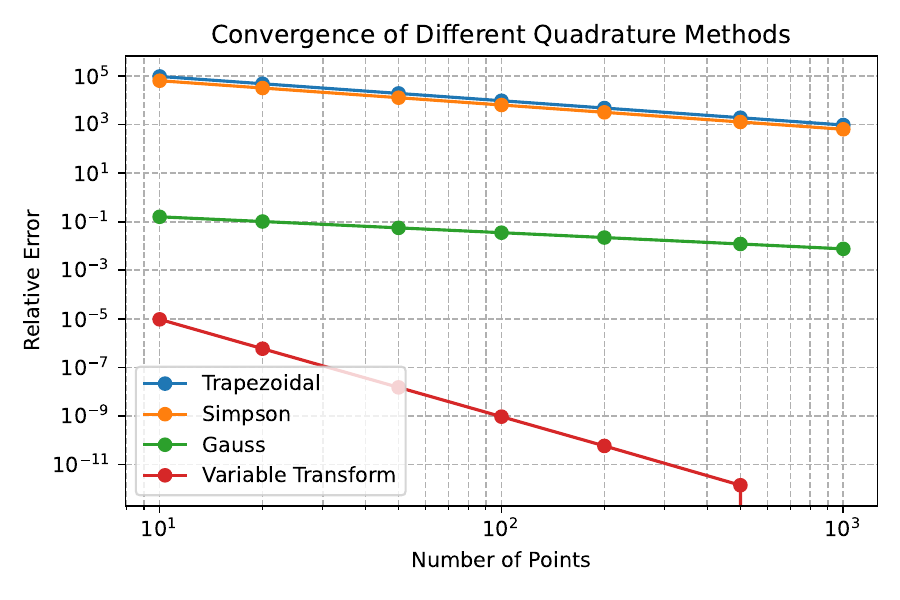}
    \caption{Convergence rate of different methods tested for integration.}
    \label{fig:comparison}
\end{figure}

After applying this transformation, most LLMs employed Gaussian quadrature rules to approximate the integral. In particular, Claude models investigated a broader set of integration techniques to assess convergence rates, including the trapezoidal rule, Simpson’s rule, direct Gaussian quadrature, and the aforementioned variable transformation. As shown in \autoref{fig:comparison}, the variable transform achieves the fastest convergence among these approaches, providing a thorough comparison of their performance.

Furthermore, in the case of Claude 3.7 Sonnet with extended reasoning, several quadrature methods are tested following the variable transformation—namely Gaussian quadrature, adaptive Simpson’s rule, and Romberg integration—offering additional insights into the efficacy of different numerical integration strategies. A detailed summary of the chosen methods and their associated errors can be found in \autoref{tab:integral}.

\begin{table}[h!]
   \centering
    \renewcommand{\arraystretch}{1.5}
    \caption{Summary of performance of different models in integration.}
    \vspace{2mm}
    \resizebox{0.9\textwidth}{!}{
    \begin{tabular}{c|c|c|c|c}
    \toprule
         & Reasoning time (s)& Method & Number of Nodes & $L_{2}$ error \\
         \hline 
         DeepSeek V3&N/A& Scipy & N/A & N/A \\ 
         \hline 
         DeepSeek R1 & 584.0&Gauss-Legendre &10 &3.74e-12 \\ 
         \hline 
         ChatGPT 4o &N/A& Gauss-Legendre & 20 &5.15e-13\\ 
         \hline  
         ChatGPT o3-mini-high &74.0& Gauss-Legendre&50 & 5.21e-14 \\ 
         \hline 
         Claude 3.7 Sonnet &N/A& Gauss-Legendre & 1000 & Machine precision\\ 
         \hline 
         Claude 3.7 Sonnet extended thinking &37.0& Gauss-Legendre & 1000 & Machine precision \\ 
         \bottomrule
    \end{tabular}}
    \label{tab:integral}
\end{table}

Overall, except for DeepSeek V3, all models can achieve comparably high performance and correctly follow the user request. They are able to identify the singularity and correctly apply the transformation before using quadrature rules. Claude models provide more comprehensive comparisons of different methods and are more sophisticated in analyzing such scientific computing tasks.

\subsection{Scientific Machine Learning}

In this section, we test the LLMs on a range of tasks in scientific machine learning, in image recognition, physics-informed machine learning, and operator learning methods. 

\subsubsection{MNIST digits prediction}

Many problems in science and engineering require analyzing image data and classification. The MNIST \cite{deng2012mnist} dataset is chosen as a test problem because it is a well-established benchmark for evaluating image recognition models. All LLMs in this experiment are asked with the following prompt: 

\texttt{Implement a CNN in tensorflow to learn the MNIST dataset. Choose appropriate architecture and hyperparameters to aim for both high accuracy and computational efficiency.}

The goal is to evaluate how the LLMs make decisions to balance the trade-off between achieving high accuracy and saving computational costs. The generated responses are summarized in \autoref{tab:MNIST}. All tests were conducted on a Quadro RTX 6000 GPU.

\begin{table}[h!]
\centering
\caption{Comparison of  models generated by DeekSeek, ChatGPT, and Claude for learning the MNIST dataset. ``Convolution($n$)" means a 2D convolution layer with $n$ channels, ``Dense($n$)" means a fully connected layer with $n$ neurons, and ``Dropout($p$)" means dropout with probability $p$.}
\vspace{2mm}
\resizebox{0.75\linewidth}{!}{%
\renewcommand{\arraystretch}{1.25}
\begin{tabular}{c|c|c|c|c|c|c}
\toprule
Model &  Reasoning&Architecture & Epochs & Time & Testing & Early \\
 &  time (s)&& & (s)& Accuracy&Stopping\\ \hline
DeekSeek V3 & 
 N/A&\begin{tabular}[c]{@{}l@{}}
Convolution(32)\\
MaxPooling\\
Convolution(64)\\
MaxPooling\\
Convolution(64)\\
Dense(64)\\
Dense(10)
\end{tabular} & 
5 & 21.30 & 98.96\% & N/A \\ \hline
DeepSeek R1 & 
 129.0&\begin{tabular}[c]{@{}l@{}}
Convolution(32)\\
MaxPooling\\
Convolution(64)\\
MaxPooling\\
Dense(128)\\
Dropout(0.5)\\
Dense(10)
\end{tabular} & 
20 & 21.23 & 99.11\% & Yes \\ \hline
ChatGPT 4o & 
 N/A&\begin{tabular}[c]{@{}l@{}}
Convolution(32)\\
MaxPooling\\
Convolution(64)\\
MaxPooling\\
Dropout(0.5)\\
Dense(128)\\
Dense(10)
\end{tabular} & 
10 & 21.85 & \textbf{99.31\%} & N/A \\ \hline
\begin{tabular}{@{}l@{}}ChatGPT \\ o3-mini-high\end{tabular} & 
 8.0&\begin{tabular}[c]{@{}l@{}}
Convolution(32)\\
Convolution(64)\\
MaxPooling\\
Dense(128)\\
Dense(10)
\end{tabular} & 
10 & 21.18 & 98.84\% & N/A \\  \hline
Claude 3.7 Sonnet & N/A & 
\begin{tabular}[c]{@{}l@{}}
Convolution(32)\\
MaxPooling\\
Convolution(64)\\
MaxPooling\\
Dense(64)\\
Dropout(0.5)\\
Dense(10)
\end{tabular} & 10 & 21.30 & 98.98\% & Yes \\ \hline
\begin{tabular}{@{}l@{}}Claude 3.7 Sonnet \\ extended thinking\end{tabular} & 38.0& 
\begin{tabular}[c]{@{}l@{}}
Convolution(32)\\
MaxPooling\\
Convolution(64)\\
MaxPooling\\
Dense(64)\\
Dropout(0.5)\\
Dense(10)
\end{tabular} & 10 & 21.33 & 99.14\% & Yes \\ 
\bottomrule
\end{tabular}
}
\label{tab:MNIST}
\end{table}

The results show that for this relatively standard and simple problem, all models can achieve comparable training time and high testing accuracy. Both Claude models and DeepSeek R1 implemented an early stopping criterion, which may stop the training of the model early according to changes in the validation loss to prevent overfitting. Among reasoning models, ChatGPT o3-mini-high had the fastest reasoning time.

\subsubsection{Physics-informed neural networks}
In this section, we will test the performance of LLM in solving the PDEs with physics-informed neural networks (PINNs)~\cite{raissi2019physics}, a framework used to solve PDEs with the power of neural networks. The Poisson equation to solve is given as 
\begin{equation*}
\begin{split}
    -\Delta u(x,y) = 1, \quad (x, y)\in [-1,1]^{2}/[0,1]^{2}
\end{split}
\end{equation*}
with zero Dirichlet boundary conditions. The question posed to all models is:

\texttt{Use PINNs to solve the 2D Poisson equation in L-shaped domain $[-1,1]^{2}/[0, 1]^{2}$ with zero boundary condition and right hand side $f(x,y) = 1$, where the coding language is Jax}.

All models generated code with errors. The issue was the use of the \texttt{grad} function in JAX to compute gradients of outputs with respect to inputs, which, when used incorrectly, can cause errors since \texttt{grad} is only valid for scalar output functions. After correcting this, all codes could work smoothly. The performance of the codes given by different models is summarized in \autoref{tab:PINNs}.

\begin{table}[h!]
    \centering
    \renewcommand{\arraystretch}{1.5}
    \caption{Network architectures, hyperparameters, and $L_2$ prediction errors for six PINN models solving the Poisson equation. The architecture [2,20,20,20,1] represents a fully connected neural network with three hidden layers, each containing 20 neurons, an input dimension of 2, and an output dimension of 1.}
    \vspace{2mm}
    \resizebox{0.99\textwidth}{!}{
    \begin{tabular}{c|c|c|c|c|c|c|c}
    \toprule
         &Reasoning time (s) & Network  & Training data & Epoch & Training time(s) & Activation function 
         & $L_{2}$ error \\
         \hline 
        DeepSeek V3 &N/A & [2, 20, 20, 20, 1]& 1000 & 5000 & 7.90 & Tanh & 929\%\\
        \hline 
        DeepSeek R1 &417.0& [2, 50, 50, 50,50, 1]& 512 & 10000& 13.80& Tanh & 54.7\%\\ 
        \hline 
        ChatGPT 4o &N/A& [2, 64, 64, 64, 64, 1] &5000 & 5000 & 15.23 & Tanh & 169\%\\ 
        \hline 
        \begin{tabular}{@{}l@{}}ChatGPT \\  o3-mini-high \end{tabular} &15.0& [2, 64, 64, 64, 1] & 200 (every epoch) & 5000 & 1740.00 & Tanh & 5.31\% \\ 
        \hline 
        Claude 3.7 Sonnet &N/A& [2, 32, 32, 32, 32,1]&2000 & 10000 & 12.10 &Tanh &\textbf{4.25\%} \\ 
        \hline 
        \begin{tabular}{@{}l@{}} Claude 3.7 Sonnet \\ extended thinking\end{tabular}  &38.0 & [2, 32, 32, 32, 1] & 5000 & 1000 & 19.00 &Tanh & 322\%\\ 
        \bottomrule
    \end{tabular}}
    \label{tab:PINNs}
\end{table}

\begin{figure}[h!]
    \centering
    \begin{overpic}[width = 0.44\textwidth]{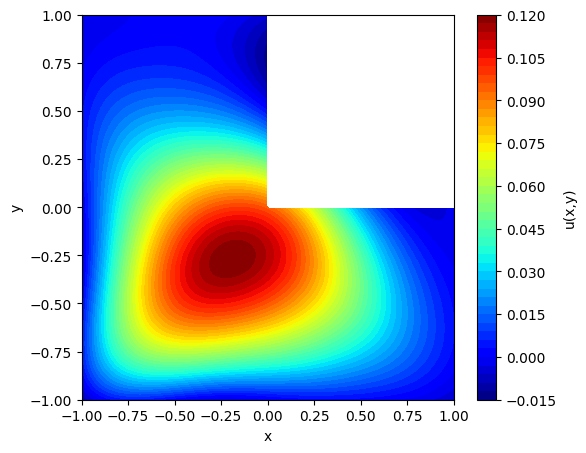}
    \put(30,76) {ChatGPT 4o}
    \end{overpic}
    \begin{overpic}[width = 0.45\textwidth]{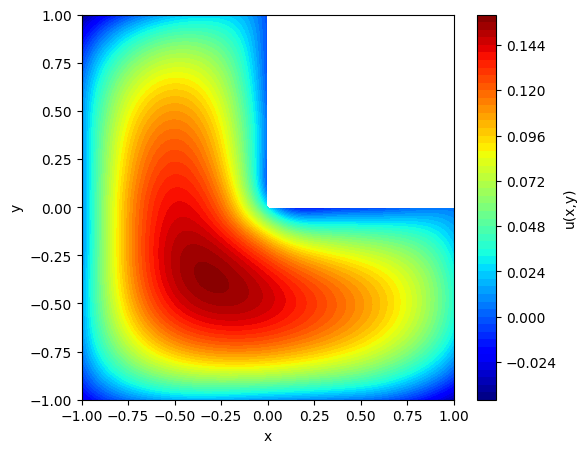}
    \put(23,76) {ChatGPT o3-mini-high}
    \end{overpic}
    \begin{overpic}[width = 0.43\textwidth]{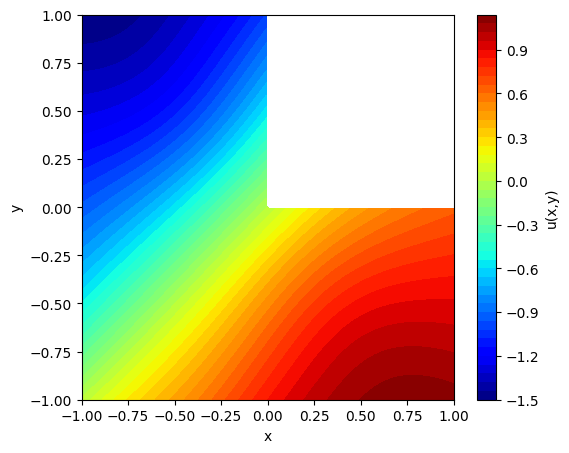}
    \put(30,78) {DeepSeek V3}
    \end{overpic}
    \begin{overpic}[width = 0.44\textwidth]{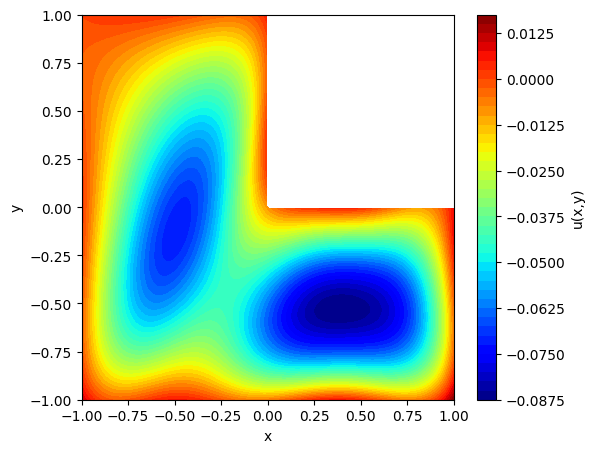}
    \put(30,75) {DeepSeek R1}
    \end{overpic}
    \begin{overpic}[width = 0.44\textwidth]{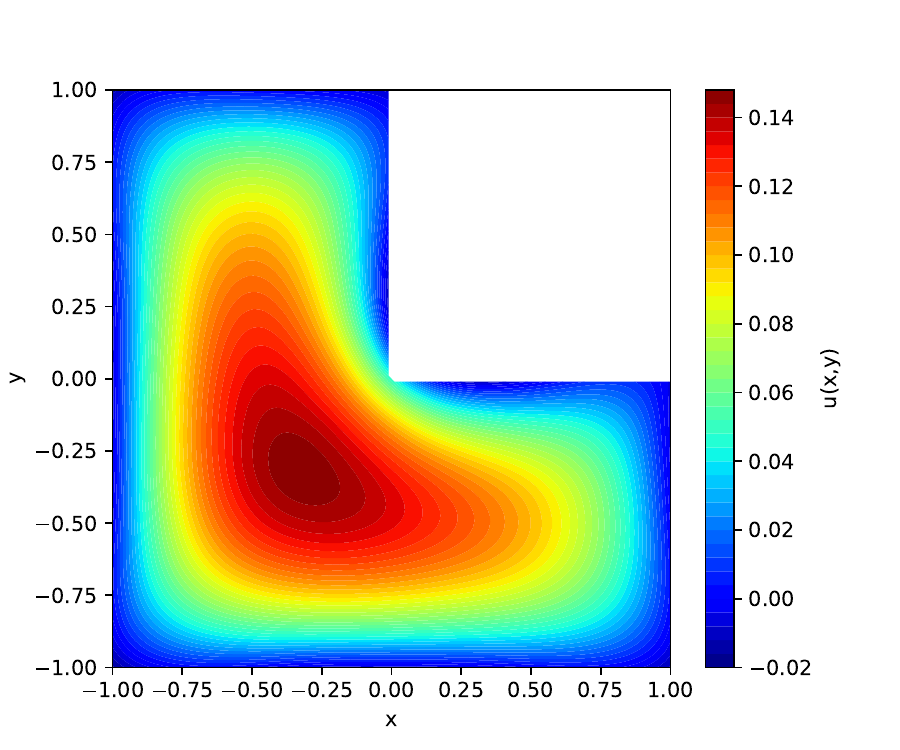}
    \put(30,75) {Claude 3.7 Sonnet}
    \end{overpic}
    \begin{overpic}[width = 0.44\textwidth]{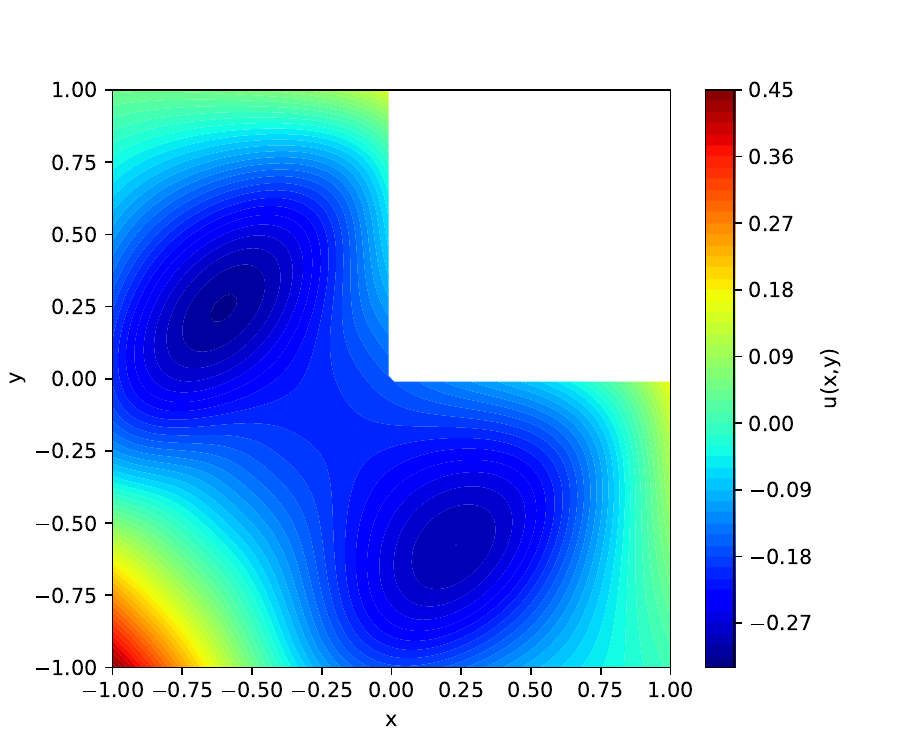}
    \hspace{-1.5cm} \put(26,75) {Claude 3.7 Sonnet extended thinking}
    \end{overpic}
    \caption{Predicted solutions from PINNs generated by different LLMs for solving the Poisson equation on an L-shaped domain.}
    \label{fig:fem_solution}
\end{figure}

Based on the results summarized in \autoref{tab:PINNs}, the full reasoning model ChatGPT o3-mini-high and the hybrid reasoning model Claude 3.7 Sonnet are able to deliver correct predictions and achieve significantly lower $L_{2}$ errors compared to the other models. These results demonstrate that, when solving the Poisson equation on an L-shaped domain, reasoning models not only capture the solution’s complex behavior more accurately but also adapt more flexibly to nontrivial modifications in the problem domain.

In contrast to non-reasoning models, which typically train within the original problem domain, DeepSeek R1 also exhibits the capability to recognize and adapt to the L-shaped boundary by generating appropriate boundary points. This adaptive strategy underlines the enhanced generalization and responsiveness afforded by incorporating reasoning capabilities into the model.

An additional noteworthy observation is the difference in reasoning times. As shown in \autoref{tab:PINNs}, while DeepSeek R1 requires a considerable 417 seconds of reasoning time, the other two reasoning‐based models respond much faster—ChatGPT o3-mini-high operates a reasoning time of only 15 seconds and Claude 3.7 Sonnet with extended thinking 38 seconds. 

Moreover, although ChatGPT o3-mini-high requires over 1000 seconds of training time, mainly due to its need to generate new samples at every epoch, the superior accuracy it achieves justifies the additional computational effort. Overall, the findings in \autoref{tab:PINNs} provide compelling evidence that PINNs generated by reasoning LLMs offer a more flexible response to the complexities of the problem domain and significantly outperform non-reasoning models.

\subsubsection{DeepONet for learning the antiderivative operator}

The Deep Operator Network (DeepONet) \cite{deeponet} is a neural operator designed to learn mappings between function spaces using data, based on the universal approximation theorem for operators \cite{chen_universal_thm_neural_operator}. Given an input function $ u: x \mapsto u(x) $ defined on a domain $ D \subset \mathbb{R}^n $, and an output function $ v: y \mapsto v(y) $ defined on a domain $ \Omega \subset \mathbb{R}^m $, the goal is to approximate the operator:
\begin{align*}
    \mathcal{G}: \mathcal{U} \ni u \mapsto v \in \mathcal{V}.
\end{align*}

The architecture of DeepONet consists of two components: a trunk network that takes the coordinates $ y \in \Omega $; and a branch network that takes a discretized version of the input function $ u $, sampled at $ m $ arbitrary sensor locations $ \{x_1, x_2, \dots, x_m\} $. The output of DeepONet is given as $v(y) = \sum_{i = 1}^r b_i(\boldsymbol{u}) t_i(y) + b_0 \approx \mathcal{G}(u)(y) $, where $ b_i $ and $ t_i $ are the outputs of the branch and trunk networks, respectively, and $ b_0 $ is a trainable bias term. 

\begin{figure}[h!]
    \centering
    \includegraphics[width=0.5\linewidth]{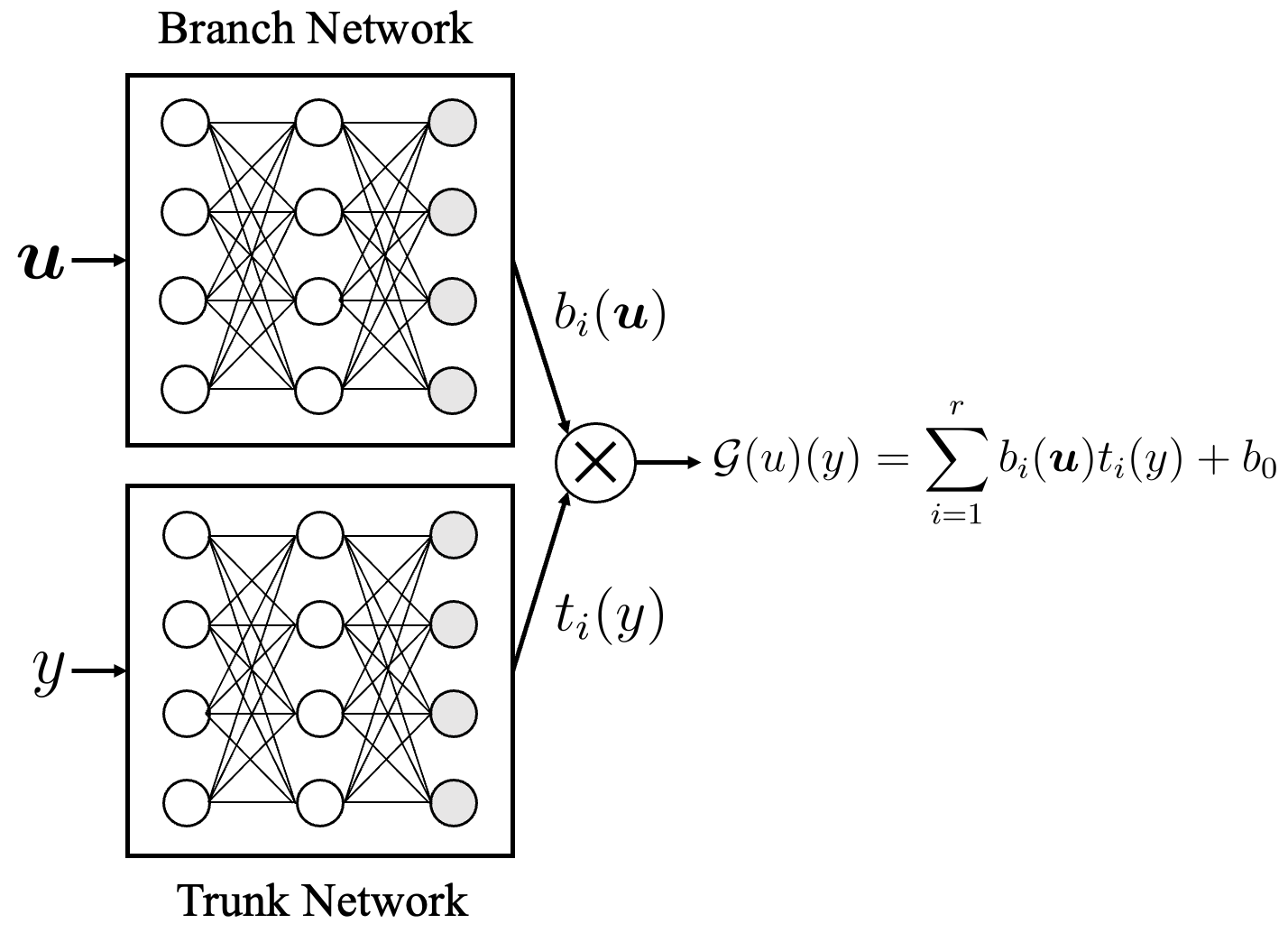}
    \caption{DeepONet architecture: The branch network encodes the input function $ u $, while the trunk network encodes the coordinates $ y $ where the output function $ v $ is evaluated.}
    \label{fig:deepONet_diagram}
\end{figure}

For a first test of the LLMs' knowledge in operator learning, we consider a task for learning the general anti-derivative operator $G(u): u(x) \mapsto s(x) = \int_0^x u(\tau) d\tau$ on a domain of $[0,1]$. All six LLMs are asked with the following prompt: 

\texttt{Implement a Deep Operator Network (DeepONet) in Tensorflow to learn the anti-derivative operator. Choose appropriate architecture and hyperparameters to aim for both high accuracy and low computational cost. Make sure that the model can generalize to a wide variety of functions. }

Note that the choice of the input function space is left deliberately ambiguous in the prompt. Therefore, the LLMs need to strategically sample from a function space, and generate the antiderivatives of these functions to use as the targets for training the DeepONet. The models' responses to the data generation task are very different:
\begin{enumerate}
    \item DeepSeek V3: Input functions are simply generated point-wise as uniformly random numbers in \texttt{numpy}. The functions are then numerically integrated using cumulative sums (\texttt{cumsum}) in \texttt{numpy}. 
    \item DeepSeek R1: Input functions are randomly generated from the following 4 types: 
    \begin{enumerate}
        \item 6th-degree polynomials $u(x) = c_0 + c_1x + \cdots c_6 x^6$ with random coefficients ranging from $[-2,2]$. 
        \item Trigonometric functions $u(x) = A \sin(\omega x) + B \cos (\omega x)$, where $A, B$ are chosen randomly from $[-2,2]$ and $\omega$ chosen randomly from $[1,5]$.
        \item Exponential functions $u(x) = A \exp(Bx)$, where $A, B$ are chosen randomly from $[-1,1]$. 
        \item A mixture of 1, 2, and 3 given as $u(x) = c_1 x^2 + c_2 \sin(\omega x) + c_3 \exp(x)$, where $c_i$ are chosen randomly from $[-1,1]$ and $\omega$ from $[1,5]$. All antiderivatives are calculated analytically. 
    \end{enumerate}
    \item ChatGPT 4o: Input functions are generated as 5-th degree polynomial without the constant term, $u(x) = c_1x + \cdots c_5 x^5$, where the coefficients $c_1$ are chosen randomly from $[-1,1]$. The antiderivatives are calculated analytically. 
    \item ChatGPT o3-mini-high: Input functions are generated as Fourier series with 3 modes $u(x) = \sum_{i=1}^3 A_i \sin(2\pi i x) + B_i \cos(2 \pi i x)$ where $A_i, B_i$ are chosen randomly from $[-1,1]$. The antiderivatives are calculated numerically using cumulative trapezoidal rule with  \texttt{scipy.integrate.cumtrapz}. 
    \item Claude 3.7 Sonnet: similar to DeepSeek R1, input functions are generated as any of the following:
    \begin{enumerate}
        \item Fourier series $u(x) = \sum_{i=1}^N A_i \sin(2\pi i x) + B_i \cos(2 \pi i x)$, where $A, B$ are chosen randomly from $[-1,1]$ and $N$ chosen randomly from $\{2,3,4,5,6\}$.
        \item Polynomial with the degree $d$ chosen randomly from $\{2,3,4,5\}$ and random coefficients from $[-1,1]$. 
        \item Trigonometric $u(x) = A \sin(B x) + C \cos(2x)$ where $A, B, C$ are chosen randomly from $[0.5, 2]$.  All antiderivatives are calculated analytically. 
    \end{enumerate}
    \item Claude 3.7 Sonnet with extended thinking: similarly, input functions are generated as any of the following:
    \begin{enumerate}
        \item Polynomial with the degree $d$ chosen randomly from $\{0, 1, 2,3,4,5\}$ and random coefficients from $[-2,2]$. 
        \item Sine function $u(x) = A \sin(B x)$ where $A, B$ are chosen randomly from $[0.5, 3]$. 
        \item Exponential function $u(x) = A \exp(Bx)$, where $A, B$ are chosen randomly from $[0.5,2]$. 
        \item Mixed function $u(x) =  A \sin( x) + B x^2$, where $A, B$ are chosen randomly from $[0.5,2]$. 
        \item Gaussian $u(x) = a \exp\left(-\left(\frac{x - \mu}{\sigma}\right)^2\right)$ where $a \sim \text{Unif}[1, 3]$, $\mu \sim \text{Unif}[-0.5, 0.5]$, and $\sigma \sim \text{Unif}[0.2, 0.5]$. The antiderivatives for Gaussian functions are calculated numerically using \texttt{scipy.integrate.quad}. 
        \item Rational function $u(x) = \frac{A}{1 + B x^2}$ where $a, b$ are chosen randomly from $[0.5,2]$. The antiderivatives for functions other than the Gaussian are calculated analytically.  
    \end{enumerate}
\end{enumerate}

We observe that Claude 3.7 Sonnet with extended thinking generates the most complex training dataset from a diverse mix of function types. Surprisingly, no model considers using Gaussian Random Fields (GRF) to generate training functions, which is the strategy used in the original DeepONet paper \cite{deeponet}. 

To proceed with the training, some bugs in the LLM-generated code were manually fixed. Empirically, we found through repeated experiments that Claude-generated code has more non-trivial bugs in this particular task compared to other models. A few major bugs include:
\begin{enumerate}
    \item DeekSeek V3's code has errors due to unmatched tensor dimensions in the dot product between branch net and trunk net outputs, which was manually corrected.
    \item ChatGPT 4o's code incorrectly sets the trunk net input dimension to 100 (the number of evaluation points) instead of 1 (the dimension of the dependent variable). The correct dimension was manually set and the tensor shapes were fixed accordingly. 
    \item Claude 3.7 Sonnet repeatedly outputs a bug where a \texttt{KerasTensor} is passed to a TensorFlow function (\texttt{tf.reshape}) instead of a Keras operation. This violated Keras' Functional API rules. This bug was fixed by wrapping the function in a custom Keras layer.
    \item Claude 3.7 Sonnet with extended thinking makes the same mistake as Claude 3.7 Sonnet. 
\end{enumerate}

The model architectures and training results are summarized in \autoref{tab:Antiderivative}. Both Claude models adopt the usage of Swish activation, defined as $\text{Swish}(x) = x \cdot \frac{1}{1 + e^{-x}}$, rather than the more traditional ReLu activation, justifying the choice by citing advantages like smoothness and non-monotonicity. The Claude models also used advanced  training techniques such as $L^2$ weight regularizer, learning rate scheduling, and dropout training. 

\begin{table}[h!]
    \centering
    \caption{Network architectures, hyperparameters, and $L_2$ prediction errors for the DeepONet models learning the antiderivative operator. The network architecture notation follows the same convention as in \autoref{tab:PINNs}.}
    \vspace{2mm}
    \resizebox{0.99\textwidth}{!}{
    \renewcommand{\arraystretch}{1.5}
    \begin{tabular}{c|c|c|c|c|c|c|c|c|c}
    \toprule
    Model & \begin{tabular}{@{}l@{}} Reasoning \\ time (s) \end{tabular}  & \begin{tabular}{@{}l@{}} Bug \\ free ? \end{tabular} &Branch Net & Trunk Net & \begin{tabular}{@{}l@{}} Num of \\ data \end{tabular} & Activation & \begin{tabular}{@{}l@{}} Batch \\ size \end{tabular} & Epochs & \begin{tabular}{@{}l@{}} Early \\ stopping ?  \end{tabular}  \\ \hline
    DeekSeek V3 & N/A &  No&[100, 128, 128, 128] & [1, 128, 128, 128] & 10,000 & ReLu & 32 & 100 & No\\ \hline
    DeepSeek R1 &  214.0&  Yes&[100,100,100,50] & [1,100,100,50] & 5,000 & ReLu & 32 & 200 & Yes \\ \hline
    ChatGPT 4o & N/A &  No&[100, 64, 64, 64, 64] & [1, 64, 64, 64, 64] & 10,000 & ReLu & 32 & 100 & No\\ \hline
    \begin{tabular}{@{}l@{}} ChatGPT \\ o3-mini-high \end{tabular}  &  11.0&  Yes&[100,100,100,100] & [1,100,100,100] & 1,000 & ReLu & 64 & 50 & No\\ \hline
    Claude 3.7 Sonnet & N/A &  No&[100, 64, 128, 64, 32]& [1, 32, 64, 32, 32]& 800& Swish& 64 & 50 & Yes \\ \hline
    \begin{tabular}{@{}l@{}} Claude 3.7 Sonnet \\ extended thinking\end{tabular}  & 79.0&  No&[100, 256, 128, 64, 50]& [1, 64, 64, 50] & 2,000& Swish & 128 & 150 & Yes \\ \bottomrule
    \end{tabular}
    }
    \label{tab:Antiderivative}
\end{table}

\begin{figure}[h!]
    \centering
    \includegraphics[width=0.75\linewidth]{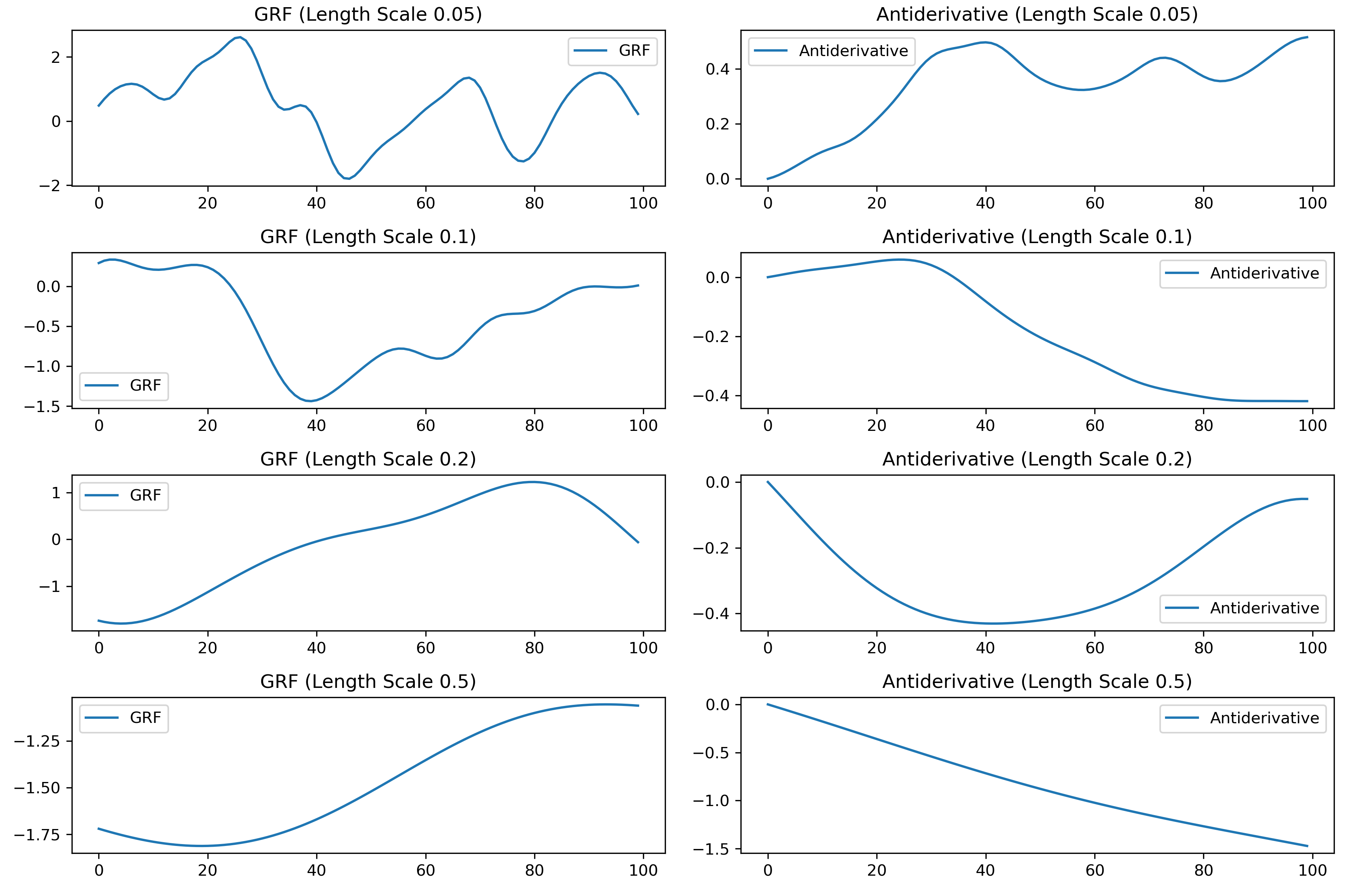}
    \caption{Testing data for DeepONets learning the antiderivative operator, generated with Gaussian Random Fields (GRF) with different length scales.}
    \label{fig:anti-derivatives}
\end{figure}

Since the models are trained on different datasets, we build a universal and general testing set to evaluate their performances. The testing set consists of functions generated by Gaussian Random Fields (GRF) with length scales  $0.05, 0.1, 0.2, 0.5$. An example of the data is shown in \autoref{fig:anti-derivatives}. The results for each model are shown in \autoref{tab:Antiderivative testing}. Functions generated with smaller length scales are more difficult to predict for most models. DeepSeek R1 gives the best result, while DeepSeek V3 fails to make meaningful predictions across all length scales. We note that the poor testing performance of Claude-generated DeepONets is due to hyperparameter choices, not training data. Using the same data from Claude 3.7 Sonnet with extended thinking, we removed the regularizer, learning rate scheduler, and dropout layers, switched to ReLU activation, and increased neurons to 100 per layer to match ChatGPT o3-mini-high. With these changes, the simplified model performed comparably to both ChatGPT models. This shows that while Claude generates a good strategy, its implementation still requires human intervention and hyperparameter tuning for optimal performance.

\begin{table}[h!]
    \centering
    \caption{Relative $L_2$ errors for DeepONet models on test datasets generated from GRF with different length scales. DeepSeek R1, with its diverse training data and error-free implementation, achieves the best results. Claude models used advanced dropout techniques, regularization, learning rate scheduling, and Swish activation, but without these, the models would have performed better and been comparable to ChatGPT's results.}
    \vspace{2mm}
    \renewcommand{\arraystretch}{1.5}
    \resizebox{0.99\textwidth}{!}{
    \begin{tabular}{c|c|c|c|c}
    \toprule
    Model & Length scale $0.5$ & Length scale $0.2$ & Length scale $0.1$ & Length scale $0.05$ \\ \hline
    DeekSeek V3 & 49.4\% & 50.6\% & 51.6\% & 47.5\% \\ \hline
    DeepSeek R1 & \textbf{0.0008}\% & \textbf{0.2}\% & \textbf{1.9}\% & \textbf{5.7}\% \\ \hline
    ChatGPT 4o & 0.05\% & 1.0\% & 3.4\% & 6.1\% \\ \hline
    ChatGPT o3-mini-high & 0.06\% & 1.0\% & 3.5\% & 6.2\% \\ \hline
    Claude 3.7 Sonnet & 0.35\% & 4.17\% & 12.69\% & 180.41\% \\ \hline
    Claude 3.7 Sonnet extended thinking & 0.22\% & 2.05\% & 5.0\% & 7.04\% \\ \bottomrule
    \end{tabular}
    }
    \label{tab:Antiderivative testing}
\end{table}

\subsubsection{DeepONet for learning the Caputo fractional derivative operator}

For a more advanced test problem, we present the task of learning the Caputo fractional derivative operator with an arbitrary fractional order. The 1D Caputo fractional derivative operator $G(u)(y,\alpha)$ is defined as 
\begin{align*}
    G(u)(y,\alpha) : \;& u(x) \mapsto s(y,\alpha) = \frac{1}{\Gamma(1-\alpha)} \int_{0}^{y} (y-\tau)^{-\alpha} u'(\tau) d\tau, \\
    & y \in [0,1], \, \alpha \in (0,1)
\end{align*}
where $\alpha$ is an arbitrary fractional order, and $u'$ denotes the first derivative of $u$. 

All LLMs are asked with the following prompt: 

\texttt{Implement a Deep Operator Network (DeepONet) in Tensorflow to learn the 1D Caputo fractional \\ derivative operator $G(u)(y,\alpha)$, which is defined as 
$$
G(u)(y,\alpha) : u(x) \mapsto s(y,\alpha) = \frac{1}{\Gamma(1-\alpha)} \int_{0}^{y} (y-\tau)^{-\alpha} u'(\tau) d\tau, \\
y \in [0,1], \, \alpha \in (0,1)
$$
where $\alpha$ is an arbitrary fractional order, and $u'$ denotes the first derivative of $u$. Function $u$ is sampled from a Gaussian Random Field with a length scale of 0.2. }

The input function space is specified in this test, but the LLMs are responsible for generating their own training and testing datasets. For data generation, the input function must be numerically differentiated and integrated. The methods used by each LLM for differentiation and integration are summarized in \autoref{tab:data_generation}. Note that while most models implemented valid numerical schemes, ChatGPT 4o and Claude 3.7 Sonnet both made a similar mistake in the integration step. Specifically, instead of computing the integral over the correct domain $[0,y]$, it mistakenly computed it over $[0,1]$. This led to a division by zero error as the term $(y-\tau)^{-\alpha}$ becomes singular when $y=\tau$. This bug was manually corrected and trapezoidal rule was adopted for integration. 

\begin{table}[h!]
    \centering
    \caption{Methods used by each LLM to compute the 1D Caputo fractional derivative for generating training and testing data.}
    \vspace{2mm}
    \renewcommand{\arraystretch}{1.5}
    \begin{tabular}{c|c|c}
        \toprule
        Model & Differentiation of $u'$ & Integration   \\ \hline
        DeepSeek V3 & First-order finite difference & Trapezoidal  rule (\texttt{numpy}) \\ \hline
        DeepSeek R1 & First-order finite difference & Left Riemann sum (from scratch) \\ \hline
        ChatGPT 4o & First-order finite difference & Incorrect bounds   \\ \hline
        ChatGPT o3-mini-high & Central difference & Trapezoidal  rule (\texttt{numpy}) \\ \hline
        Claude 3.7 Sonnet  & Central difference   & Incorrect bounds \\ \hline
        Claude 3.7 Sonnet extended thinking & Central difference  & Trapezoidal  rule (\texttt{numpy}) \\ \bottomrule
    \end{tabular}
    \label{tab:data_generation}
\end{table}

We are specifically interested in seeing if the LLMs know how to correctly encode the fractional order $\alpha$ into the DeepONet, as the fractional order $\alpha$ does not usually appear in other DeepONet applications and the LLMs' knowledge in this particular task might be limited. In the original DeepONet paper \cite{deeponet}, the order $\alpha$ along with the evaluation coordinates $y$ are put together into the trunk network. Notably, all six LLMs correctly placed $\alpha$ into the trunk. However, across repeated experiments, all DeepSeek and ChatGPT models consistently made the same mistake in formulating the input tensor. Specifically, they incorrectly paired each evaluation point $y$ with a different $\alpha$ and structured the input as follows:  
\begin{align*}
    \text{Trunk net input} = \begin{bmatrix} 
    y_1 & \alpha_1 \\ 
    y_2 & \alpha_2 \\ 
    \vdots & \vdots \\ 
    y_N & \alpha_N 
    \end{bmatrix}
\end{align*}
However, for each fractional derivative order $\alpha$, there should be multiple evaluation points $\{y_1, y_2, \ldots, y_n\}$ at which the output function is computed. Using distinct pairs of $(y_i, \alpha_i)$ reduces to evaluating the fractional derivative at a single point $y_i$ per $\alpha_i$ and therefore prevents the DeepONet from learning a complete representation of the derivative across the domain on $[0,1]$. As a result, DeepONets trained on the original DeepSeek or ChatGPT-generated implementation, without human intervention, yield more than $150\%$ $l^2$ relative error when tested. Since comparison under such an incorrect formulation does not provide meaningful insights, we manually corrected the code by DeepSeek and ChatGPT to ensure that the spatial input to the trunk network, $y$, is evaluated at 100 distinct points per the fractional order $\alpha$, which is randomly chosen from the interval $[0.1, 0.9]$.

Both Claude models correctly formulated the trunk net input in all experiments. However, they consistently tiled the branch net and trunk net input tensors to match their first dimension before passing them into the network. This ensured that their dot product could be directly computed after they passed through the two branches, but this tiling was unnecessary and expensive, as the dot product could be implemented easily with \texttt{einsum} or matrix multiplication in TensorFlow. The excessive tiling led to very large tensor sizes and slowed down training. 

Additionally, DeepSeek V3 mistakenly trained the DeepONet using only a fixed fractional order $\alpha = 0.5$, despite the user request specifying $\alpha \in (0,1)$ as arbitrary. Similarly, Claude 3.7 Sonnet with extended thinking used five fixed fractional orders, $\alpha = 0.17, 0.25, 0.34, 0.65, 0.73$, across all 900 training functions, resulting in a total training sample size of 4500. Fixing the fractional order in training data reduced the DeepONet's ability to generalize to unseen fractional orders during testing. The correct approach is to sample $\alpha$ randomly for each training sample. However, these mistakes were not manually corrected as they are part of the LLMs' decision making.  

Finally, we train and test all DeepONets on GRF-generated functions with length scale $0.2$ and random orders $\alpha \in [0.1, 0.9]$. The results, along with model architecture and hyperparameters, are reported in \autoref{tab:DeepONet_fractional}.  

\begin{table}[h!]
    \centering
    \caption{Network architectures, hyperparameters, and $L_2$ prediction errors for the DeepONet models learning the 1D Caputo fractional derivative. The notation follows \autoref{tab:PINNs}. The testing $L_2$ errors reported in this table reflect corrections made after manually fixing bugs in the original code. Claude 3.7 Sonnet achieves the best result but is slower due to inefficient tensor manipulation in the code. The performances of DeepSeek V3 and Claude 3.7 Sonnet with extended thinking are hindered by using a fixing fractional order $\alpha$ across different training samples. }
    \vspace{2mm}
    \resizebox{0.99\textwidth}{!}{%
    \renewcommand{\arraystretch}{1.5}
    \begin{tabular}{c|c|c|c|c|c|c|c|c|c}
    \toprule
    Model &  \begin{tabular}{@{}c@{}}Reasoning \\ Time (s) \end{tabular} & \begin{tabular}{@{}c@{}}Bug \\ free ?\end{tabular} &Branch Net & Trunk Net & Epochs & \begin{tabular}{@{}c@{}}Num of \\ train data\end{tabular} &  Activation&Training Time (s) & Testing $L_2$ Error \\ \hline
    DeepSeek V3 & N/A &  No&[100, 128, 128, 128] & [2, 128, 128, 128] & 100 &  1000&  ReLu&11.87 & 177.19\%\\ \hline
    DeepSeek R1 & 426.0&  No&[100, 128, 128, 128] & [2, 128, 128, 128] & 50 &  1000&  ReLu&6.58 & 26.82\%\\ \hline
    ChatGPT 4o & N/A &  No&[100, 50, 50, 50, 100] & [2, 50, 50, 50, 100] & 100 &  1000&  ReLu&13.54 & 19.49\%\\ \hline
    \begin{tabular}{@{}c@{}}ChatGPT \\ o3-mini-high\end{tabular}  & 23.0&  No&[100, 128, 128, 100] & [2, 128, 128, 100] & 100 &  1000&  ReLu&10.91 & 18.41\%\\ \hline
    Claude 3.7 Sonnet & N/A &  No&[100, 128, 128, 40]& [2, 128, 128, 40]& 200&  900&  ReLu&41.74 & \textbf{5.59}\% \\ \hline
    \begin{tabular}{@{}c@{}}Claude 3.7 Sonnet \\ extended thinking\end{tabular} & 109.0&  No&[100, 128, 128, 128, 50]& [2, 128, 128, 128, 50]& 50&  4500&  ReLu&1943.14 & 16.24\% \\ 
    \bottomrule
    \end{tabular}%
    }
    \label{tab:DeepONet_fractional}
\end{table}

The testing $L_2$ errors reported in this table reflect corrections made after manually fixing bugs in the original code. Note that the testing error for DeepSeek V3 is very high, as it is trained on a single fractional order and thus fails to generalize. Claude 3.7 Sonnet with extended thinking also suffers from the same mistake, but its larger sample size and choice of five different $\alpha$'s contributes to better performance.  Overall, the hybrid reasoning model Claude 3.7 Sonnet achieves the best result but is slower due to inefficient tensor manipulation in the code. Additionally, similar to our previous experiment results, most models are only trained for no more than 100 epochs, which is insufficient for proper convergence. In fact, during our experiments, simply increasing the number of epochs to 1000 reduces the relative error to single digits, with exactly the same model.

\section{Summary}
In this study, we evaluated and compared the performance of recently released large language models from DeepSeek (DeepSeek V3, DeepSeek R1), OpenAI (ChatGPT 4o, ChatGPT o3-mini-high), and Anthropic (Claude 3.7 Sonnet, Claude 3.7 Sonnet with extended thinking) for diverse tasks in scientific computing and scientific machine learning. We designed challenging problems that require advanced mathematical reasoning and domain-specific knowledge, including numerical integration, finite difference methods (FDM), finite element methods (FEM), and scientific machine learning tasks such as image recognition, physics-informed neural networks (PINNs), and Deep Operator Networks (DeepONet). Our focus was on assessing the models’ ability to select appropriate numerical methods or neural network architecture and implement them correctly in Python. The results show that reasoning-optimized models, DeepSeek R1, ChatGPT o3-mini-high, and Claude with extended thinking, consistently performed better in recognizing the nature of the problem and making their decisions accordingly. In fact, many of their choices are similar to what we would do in solving these problems both for scientific computing and scientific  machine learning. In contrast, general-purpose models, DeepSeek V3 and ChatGPT 4o, sometimes fail to account for specific properties of the problem (such as stiffness) or instructions from the user (such as to implement from scratch), consequently generating incorrect solutions.

Our study highlights the growing practicality of using LLMs in scientific research, as well as the increasing competition between DeepSeek, OpenAI, and Anthropic, as all front-running developers continue to refine their models for tasks in mathematics, coding, and scientific computing. Our findings also expose the limitations of these state-of-the-art LLMs, demonstrated by their ambiguous or incorrect responses, which could confuse a human researcher unfamiliar with the subject. Our findings highlight the need for continued improvements in LLMs for scientific problem-solving. Future work should explore additional benchmarking methods and assess LLMs on more complex, real-world computational challenges, where a cascade of decisions is required at the different stages of a project. By doing so, researchers can make more informed decisions on when and how to use LLM assistants effectively and responsibly in their daily work.

\section*{Acknowledgments}
This work was supported by the ONR Vannevar Bush Faculty Fellowship (N00014-22-1-2795). We would like to thank the members of the Crunch Group at Brown University for their suggestions and insights.

\bibliographystyle{unsrt}  
\bibliography{ref}

\end{document}